\documentclass[twoside,runningheads,10pt]{llncs}
\newcommand{\defaultLlncsTextWidth}[0]{12.2cm}
\newcommand{\textWidthForPeusdocode}[0]{15cm}
\usepackage{geometry}
\newcommand{\setHMarginsBeforePeusdocode}[0]{\newgeometry{textwidth=\textWidthForPeusdocode,hratio=1:1}}
\newcommand{\restoreHMarginsAfterPeusdocode}[0]{\newgeometry{textwidth=\defaultLlncsTextWidth,hratio=1:1}}

\AtBeginDocument{%
  \providecommand\BibTeX{{%
    \normalfont B\kern-0.5em{\scshape i\kern-0.25em b}\kern-0.8em\TeX}}}

\usepackage[disable]{todonotes}

\usepackage[inline]{enumitem}

\usepackage[breaklinks=true]{hyperref}
\usepackage{breakcites}
\usepackage{bbm}
\usepackage{amsmath}
\usepackage{amssymb}
\usepackage{booktabs}
\usepackage{tabularx}
\usepackage[capitalise]{cleveref} 
\usepackage{multirow}
\usepackage[caption=false]{subfig}
\usepackage{listings}
\usepackage{parnotes}
\usepackage{seqsplit}

\usepackage[utf8]{inputenc}
\usepackage[LAE,LFE,T1]{fontenc}
\usepackage[farsi,english]{babel}
\definecolor{codegreen}{rgb}{0,0.6,0}
\definecolor{codegray}{rgb}{0.5,0.5,0.5}
\definecolor{codepurple}{rgb}{0.58,0,0.82}
\definecolor{backcolour}{rgb}{0.95,0.95,0.92}
 
\lstdefinestyle{mystyle}{
    backgroundcolor=\color{backcolour},   
    commentstyle=\color{codegreen},
    keywordstyle=\color{magenta},
    numberstyle=\tiny\color{codegray},
    stringstyle=\color{codepurple},
    basicstyle=\tiny,
    breakatwhitespace=false,         
    breaklines=true,                 
    captionpos=b,                    
    keepspaces=true,                 
    numbers=left,                    
    numbersep=5pt,                  
    showspaces=false,                
    showstringspaces=false,
    showtabs=false,                  
    tabsize=2
}
 
\usepackage[linesnumbered,lined,boxed,commentsnumbered]{algorithm2e}

\crefname{lstlisting}{listing}{listings}
\Crefname{lstlisting}{Listing}{Listings}

\usepackage{bbm}
\usepackage{mathrsfs}

\usepackage{amsmath}
\usepackage{amssymb}

\newcounter{algolineAcrossEnvs}
\newcommand\startThisEnvCounter{\setcounter{AlgoLine}{\thealgolineAcrossEnvs}}
\newcommand\updateGlobalAlgoLineCount{\setcounter{algolineAcrossEnvs}{\theAlgoLine}}

\newcommand\partialFunc{\rightharpoonup}
\newcommand\setOfAllBoxes{\mathscr{B}}

\newcommand{\AID}{AID}
\newcommand{\AIDs}{AIDs}

\newcommand{\AIDsFull}[0]{abstract interpretation domains}

\newcommand{\CEGAR}[0]{CEGAR}
\newcommand{\CEGARFull}[0]{Counter Example Guided Abstraction Refinement}

\newcommand{\dateUpdatePostVMCAITwentyTwentyTwo}{February $\text{15}^{th}$, 2022}

\usepackage{wrapfig}

\newcommand{\FanoosState}{state\xspace}
\newcommand{\FanoosStates}{states\xspace}
\usepackage{mathtools}
\newcommand{\closedint}[2]{\ensuremath{[#1,#2]}}
\newcommand{\firstn}[1]{\ensuremath{[#1]}}
\newcommand{\ithval}[2]{\ensuremath{\{#1\}_{#2}}}
\newcommand{\indfun}[1]{\ensuremath{\mathbbm{1}(#1)}}

\newcommand{\vecArrow}[1]{\overrightarrow{#1}} %

\begin{document}

\title{Fanoos: Multi-Resolution, Multi-Strength, Interactive Explanations for Learned Systems\thanks{This material is based upon work supported by the United States Air Force and DARPA under Contract No. FA8750-18-C-0092. Any opinions, findings and conclusions or recommendations expressed in this material are those of the author(s) and do not necessarily reflect the views of the United States Air Force and DARPA.}}

\titlerunning{Fanoos: Explanations for Learned Systems}

\author{David Bayani\orcidID{0000-0001-5811-6792} \and Stefan Mitsch\orcidID{0000-0002-3194-9759}}

\authorrunning{David Bayani, Stefan Mitsch}
\institute{Computer Science Department\\Carnegie Mellon University, Pittsburgh PA 15213, USA\\
\email{dcbayani@alumni.cmu.edu, smitsch@cs.cmu.edu}
}
\maketitle

\begin{abstract}
Machine learning is becoming increasingly important to control the behavior of safety and financially critical components in sophisticated environments, where the inability to understand learned components in general, and neural nets in particular, poses serious obstacles to their adoption.
Explainability and interpretability methods for learned systems 
have gained considerable academic attention, but the focus of 
current approaches
on only one aspect of explanation, at a fixed level of abstraction, and limited if any formal guarantees, prevents those explanations from being digestible by the relevant stakeholders (e.g., end users, certification authorities, engineers) with their diverse backgrounds and 
situation-specific needs.
We introduce Fanoos, a framework for combining formal verification techniques, heuristic search, and user interaction to explore explanations at the desired level of granularity and fidelity.
We demonstrate the ability of Fanoos to produce and adjust the abstractness of explanations in response to user requests on a learned controller for an inverted double pendulum and on a learned CPU usage model.

\end{abstract}

\section{Problem Overview}
\label{sec:q1_problemOverview}
Explainability and safety 
in machine learning 
(ML) are a subject of increasing academic and public concern.
As ML continues to grow in success and adoption by wide-ranging industries, 
the impact of these algorithms' behavior on people's lives is becoming highly non-trivial.
 Unfortunately, many of the most performant contemporary ML algorithms---neural networks (NNs)
 in particular---are widely considered
black-boxes, with the method by which they perform their duties not being amenable to 
direct human comprehension. The inability to understand learned components as thoroughly
 as more traditional software
poses serious obstacles to their adoption 
\cite{anjomshoae2019explainable,adadi2018peeking,chakraborti2019explicability,guidotti2019survey,DBLP:journals/corr/abs-1805-09944,garcia2015comprehensive,yasmin2013neural,miller2017explainable} 
due to safety concerns, difficult debugging and maintenance, and 
explicit legal requirements (e.g., the ``right to an
explanation'' legislation \cite{eugdpr2016}
adopted by the European Union).
Symbiotic human-machine interactions can lead
to safer and more robust agents, but this task requires  
effective and versatile communication \cite{veloso2015cobots,rosenthal2010effective}.

Interpretability of learned systems has been studied in the context of computer science intermittently since at least the late 1980s, particularly in the area of rule extraction (e.g., \cite{KBSSurvey1995}),   
adaptive/non-linear control analysis (e.g., \cite{david1988design}),
various rule-learning paradigms~(e.g., inductive logic programming\cite{muggleton1999inductive}, association rule learning  \cite{agrawal1993mining}, and its predecessors \cite{piatetskyknowledge,fayyad1996advances}), and formal analysis (e.g., 
\cite{clarke2003verification,wen1996towards,wen1997verifying,walter286277,kearfott1996interval,moore1966interval}).
Notwithstanding this long history, main-stream attention has risen only recently
due to increased impact on daily life of opaque AI
\cite{adadi2018peeking} with novel initiatives focused on the problem domain, 
e.g. programs by DARPA \cite{gunning2017explainable,gunning2019explainable,neemaassuredslides} as well as workshops in IJCAI (\cite{aha2017ijcai}) and ICAPS (\cite{icapsxaip2018}).

Despite this attention, however, most explanatory systems developed for ML
are hard-coded to provide a single type 
of explanation with descriptions at a certain fixed level of abstraction and
a fixed type of guarantee about the system behavior, if any (most often, formal guarentees are missing entirely). %
This 
not only prevents the explanations generated from being digestible by multiple audiences 
(the end-users, the intermediate engineers who are non-experts in the ML component, and the 
ML-engineers for instance)\footnote{
Outside of our efforts, this was for instance highlighted
by a recent taxonomy 
presented in \cite{DBLP:journals/corr/abs-1909-03012}. %
}
but in fact limits the use by any single audience since the levels
of abstraction and formal guarantees needed are situation and goal specific, not just 
a function of the recipient's background. 
When using a microscope, one varies between low and high magnification in order to find 
what they are looking for and explore samples; these same capabilities are desirable for
XAI for much the same reasons.

For example,
most consumers of autonomous vehicles may prefer to
ask general questions --- for instance, ``What do you do when you detect a person in front of you?'' ---
and receive a break-down of qualitatively different behaviors for different
situations, such as braking when traveling slowly enough, and doing a sharp swerve when
traveling too fast to brake. 
An engineer checking actuator compliance, however, might require greater details, opting to specify precise parameters of the scene
and preferring that the car report exact motor commands; the context of use and the audience
determine which level of abstraction is best, and supporting multiple types of
abstractions in turn supports more use-cases and audiences.
Further, the explanations for such
 a component need to range from formal guarantees to rough tendencies---it may be critical
to formally
 guarantee that the car will always avoid collisions, while it might be sufficient 
that it usually (but perhaps not always) drives slowly when its battery is low.

The divide between formal and probabilistic explanations also relates to events that
are imaginable versus events that may actually occur; formal methods may check every point in a
space for conformance to a condition, but if bad behavior only occurs on measure-zero sets,
the system would be safe while not being provably so in formalizations lacking knowledge
of statistics (e.g., if some criteria demands that a car keep distance ${>}10\text{\,cm}$ from obstacles, 
formally we can get arbitrarily close but not equal; in practice, the difference with
 ${\geq}10\text{\,cm}$ might be irrelevant).
Explainable ML systems should enable these sorts of search and smooth variation in need---but at the moment they do not in general.

To address these needs, we introduce Fanoos,\footnote{``Fanoos'' (\FR{فانوس})
means lantern in Farsi. Our approach shines a light on
black-box AI. Source code can be found at \url{https://github.com/DBay-ani/Fanoos} as well as \cite{david_bayani_2021_5513079}.} an algorithm blending a diverse array of
technologies to interactively provide explanations at varying levels of abstraction and fidelity (i.e., probabilistic versus formal guarentees) to meet user's needs.
Our algorithm is applicable to currently ubiquitous ML methods---such as feed-forward neural networks (FFNNs) and high-dimensional polynomial regressions.
Fanoos offers the following 
combination of capabilities, which are our contributions:
\begin{itemize}
\item Explanations that can vary in abstraction level in order to suit user's situational needs.
\item Interactivity that allows users to query the learned system they want to understand, and receive
    explanations 
    capable of characterizing the
    input requirements, output behavior, or the combination of the the two.
\item Explanations that can either be formally sound or probabilistic based on the user's choice.
    Formal soundness 
    is a capability missing from the vast majority of XAI systems focused on ML, and leveraging
    verification techniques for ML-related XAI has been underexplored.
\end{itemize}

\section{The Methodology of Fanoos}
\label{sec:q3_methodology}

Fanoos is an interactive system that allows users to pose a variety of questions grounded in a domain specification (e.g., asking what environmental
conditions cause a robot to swerve left),
receive replies from the system, and request that explanations be made more or less abstract.
An illustration of the process and component interactions  can be found in \cref{sec:appendix:diagram}, with a fuller example of interaction and discussion of the UI located in \cref{sec:sup:extendedIO}. 
A more detailed technical overview of the implementation can be found in \cref{sec:appendix:moretechdetails}.
Crucially, Fanoos provides explanations of high fidelity\footnote{Since Fanoos is a decompositional approach; see \cref{sec:q2_relatedWork}.} while considering whether the explanation should be formally sound or probabilistically
reasonable (which removes the ``noise'' incurred by measure-zero sets that can plague formal descriptions). 
To this end, we combine techniques from formal verification,
interactive systems, and heuristic search over knowledge domains when responding to user questions and requests.
Here, we do not focus on the information presentation aesthetics so much as ensuring that the proper information can be produced (see \cref{subsec:FanoosAndPrettyUI}).

\subsection{Knowledge Domains and User Questions}

\begin{table*}[h]
\caption{Description of questions that can be posed to Fanoos.}
\label{tab:questionDescriptions}
\hspace{-3cm}
\begin{tabularx}{1.5\textwidth}{p{2cm}@{\hspace{1em}}p{4cm}!{\color{lightgray} \vrule width 1pt}p{3.3cm}!{\color{lightgray} \vrule width 1pt}p{0.5cm}!{\color{lightgray} \vrule width 1pt }p{1.3cm}@{\hspace{0em}}X}
\toprule	
	    Type $q_t$ 
            & Description 
            & \multicolumn{3}{l}{Question content $q_c$}
            & Example\\
  \cmidrule(lr){3-5}          
  & & \multicolumn{2}{l}{accepts \hfill illum.} & restrictions\\            
	\midrule
            When Do You
            & Tell user all sets (formal consideration of all cases) in the input space $S_I$ that have the potential to cause $q_c$ 
            & Subset $s$ of $S_O$ s.t. there exists a member of $s$ that causes $q_c$ to be true. Found with SAT solver. 
            & $S_I$ 
            & Cannot contain variables from $S_I$. 
            & 
            \multirow{1}{*}[-2ex]{
						$\textsf{when\_do\_you} \underbrace{\textsf{move\_at\_high\_speed?}}_{\text{Predicate } p \in D}$}
            \\
\midrule
            What Do You Do When
            & Tell user all possible learner responses in the collection of input states that $q_c$ accepts 
            & Subset $s$ of $S_I$ s.t. there exists a member of $s$ that causes $q_c$ to be true. Found with SAT solver. 
            & $S_O$ 
            & Cannot contain variables from $S_O$. 
            &
            \multirow{1}{*}[-2ex]{ 
						$\begin{aligned}            
            \textsf{what\_do\_you}&\textsf{\_do\_when}\\[-1ex] 
            \textit{and}(~&\textsf{close\_to\_target\_orientation},\\[-1ex]
            &\textsf{close\_to\_target\_position}~)?
						\end{aligned}$
						}
            \\
	\hline
            What~are the Circumstan\-ces in Which
            & Tell user information about what input-output pairs occur in the subset of input-outputs accepted by $q_c$ 
            & Subset $s$ of $S_{IO}$ s.t. there exists a member of $s$ that causes $q_c$ to be true. Found with SAT solver. 
            & $S_{IO}$ 
            & None 
            & 
            \multirow{1}{*}[-1ex]{
						$\begin{aligned}            
            \textsf{what\_are}&\textsf{\_the\_circumstances\_in\_which}\\[-1ex] 						&\textit{and}(~\textsf{close\_to\_target\_position},\\[-1ex]
            &\phantom{\textit{and}(~}\textsf{steer\_to\_right}~)\\[-1ex] 
            &\textit{or}~\textsf{move\_at\_low\_speed}?
            \end{aligned}$
            }
\\
\midrule
\ldots Usually 
& Statistical tendency. Avoids measure-zero sets that are unlikely seen in practice.
& Subset over which $q_c$ was found to be true at least once via statistical sampling. Discussion in \cref{subsec:moreAboutNotionOfUssuallyTrue}.
&
&
&
$\begin{aligned}
\textsf{when\_do}&\textsf{\_you\_usually}\\[-1ex] 
& \textsf{move\_at\_low\_speed}~\textit{or}~\textsf{steer\_to\_left}?
\end{aligned}$
$\begin{aligned}
  \textsf{what\_do}&\textsf{\_you\_usually\_do\_when}\\[-1ex] 
  &\textsf{moving\_toward\_target\_position}?
  \end{aligned}$
$\begin{aligned}
  \textsf{what\_are}&\textsf{\_the\_usual\_circumstances\_in\_which}\\[-1ex] &\textit{and}(~\textsf{close\_to\_target\_position},\\[-1ex] &\phantom{\textit{and}(~}\textsf{steer\_close\_to\_center}~)?
  \end{aligned}$
\\
\bottomrule
\end{tabularx}
\end{table*}

 In the following discussion, let $L$ be the 
learned system under analysis (which we will assume is piece-wise continuous),
 $q$ be the question posed by the user, $S_I$ be the 
(bounded) input space to $L$, and $S_O$ be the output space for $L$, $S_{IO}{=}S_I \cup S_O$ be the 
joint of the input and output space\footnote{Subscripts $I$ for input, $O$ for output, etc., are simply symbols, not any sort
of richer mathematical object.}, and $r$ be the response given
by the system. In order to formulate question $q$ and response $r$, a library listing basic domain information ($D$)
is provided to Fanoos; $D$ lists what $S_I$ and $S_O$ are 
and provides a set of predicates, $P$, expressed over the domain symbols in $S_{IO}$, i.e., for all $p \in P$,
the free variables $FV(p)$ are chosen from the variable names $V(S_{IO})$, that is $~\textit{FV}(p) \subseteq \textit{V}(S_{IO})$.

\begin{lstlisting}[
  caption={Question to illuminate input space $S_I$},
  label={lst:idpquestion},
	linerange={1-1},
	basicstyle=\footnotesize,
	breakatwhitespace=true,
	breaklines=true,
  postbreak=\mbox{{$\hookrightarrow$}\space}
]
(Fanoos) when_do_you_usually and(outputtorque_low , statevalueestimate_high )?
\end{lstlisting}

For queries that formally guarantee behavior (see the first three rows in 
\cref{tab:questionDescriptions}), we require that the relevant predicates in $P$ can  
expose their internals as first-order formulas; this enables us to guarantee they are
 satisfied
over all members of a given set\footnote{The box abstractions we introduce in a
moment to be more precise.} via typical SAT solvers (such as 
Z3~\cite{DeMoura:2008:ZES:1792734.1792766}). The other query types require only 
being able to evaluate question $q$ on a variable assignment 
provided. The members of $P$ can be generated in a variety of ways, e.g., by forming most 
predicates through procedural generation and then using a few hand-tailored predicates to capture
particular cases.\footnote{For example, operational definitions of
``high'', ``low'', etc., might be derived from sample data by setting thresholds on quantile values---e.g., 90\% or higher might be considered ``high''.} Further, since our
predicates are grounded, they have the potential to be generated from example or demonstration (e.g., a discussed in \cref{subsec:sup:predicategen}).
That is, $P$ is user-extensible and may be generated by automated, semi-automated or manual means.

\subsection{Reachability Analysis of the Learned System, $L$}

Having established what knowledge Fanoos is given, we proceed to explain our process.
First, users select a question type $q_t$ and the content of the 
question $q_c$ to query the system. That is, $q=(q_t, q_c)$, where $q_t$ is a member of the first column of 
\cref{tab:questionDescriptions} and $q_c$ is a sentence in disjunctive normal form (DNF)
over a subset of $P$ that obeys the restrictions 
listed in \cref{tab:questionDescriptions}. To ease discussion, we will refer to variables and
sets of variable assignments that $q$ accepts ($\textit{AC}_q$) and those that $p$ illuminates ($\textit{IL}_q$), with the
intuition being that the user wants to know what configuration of illuminated variables result
 in (or result from) the variable configurations accepted by $q_c$; see \cref{tab:questionDescriptions} for example queries.
In other words, when a user asks a question, Fanoos answers by describing a collection of situations
that necessarily include those related to the user's question; this answer is conservative in that it may
include additional situations, but never intentionally excludes cases.

\begin{figure}[htb]
\includegraphics[width=(\linewidth)]{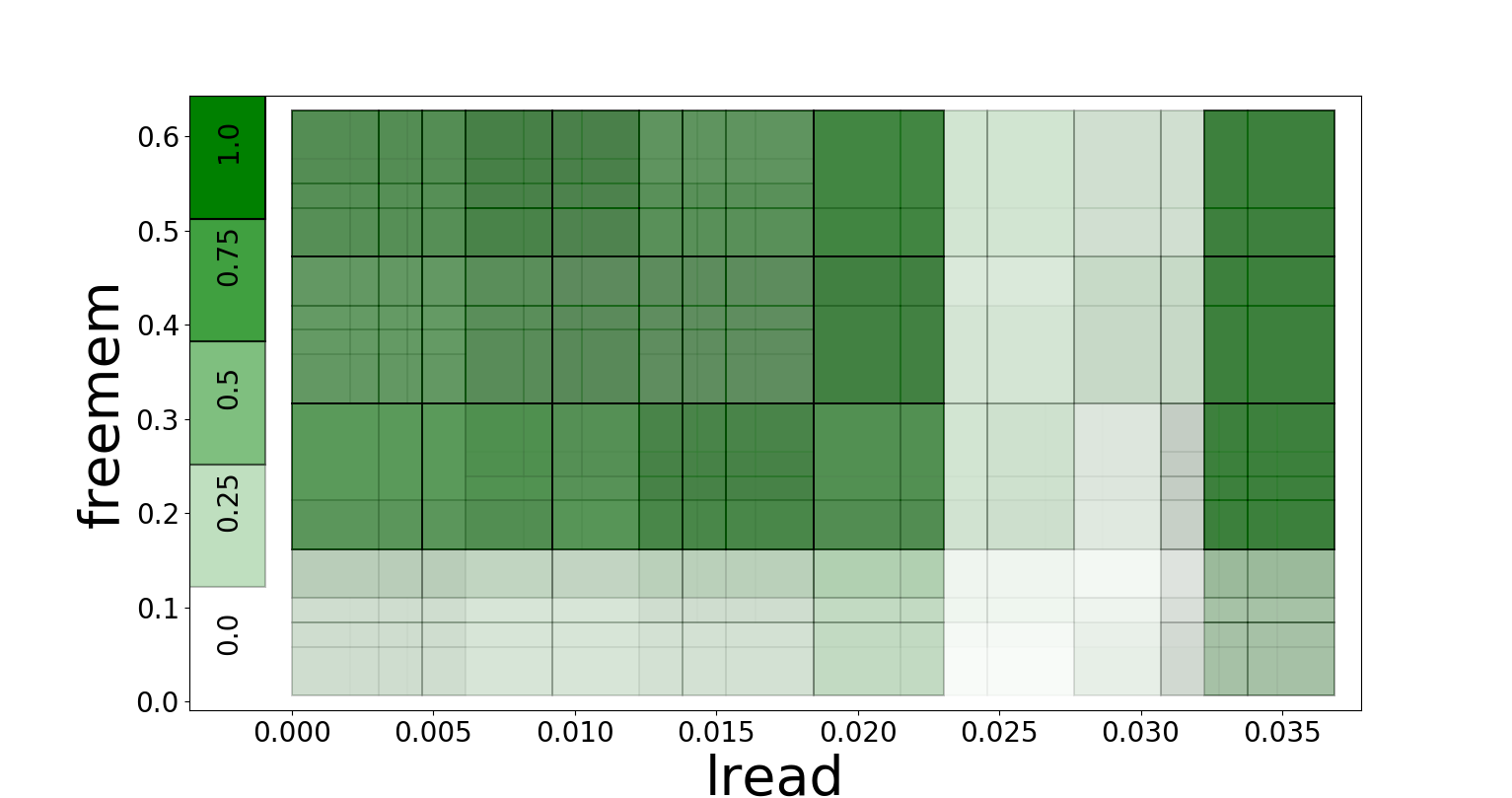}
\centering
        \caption{Example of reachability results. Shown is a 2-D projection of 5-D input-boxes
	which were selected for retention since their corresponding output-boxes satisfied the user query. 
	Darker areas had boxes with greater volume along the axes not shown; the volumes were normalized in 
	respect to the input-space bounding-box volume over the non-visible axes. A scale is along the y-axis.}
        \label{fig:reachabilityExample}
\end{figure}

With question $q$ provided, we analyze the learned system $L$ to find subsets in the inputs $S_I$
and outputs $S_O$ that agree with configuration $q_c$ and may over-approximate the behavior 
of $L$. Specifically, we adopt an approach inspired by \CEGAR\ \cite{clarke2000counterexample,clarke2003verification} with
boxes (hyper-cubes) as abstractions and a random choice between a bisection
or trisection along the longest normalized axis as the refinement process
to find the collect of box tuples, B, specified below:
\begin{multline*}
B{=}\{(B_I^{(i)}, B_{O}^{(i)}) \in \text{BX}(S_I){\times} \text{BX}(S_O)~| B_{O}^{(i)} \supseteq L(B_I^{(i)}) \\
\land ( \exists\, (c,d) \in \text{T}. (\textit{AC}_{q}(B_{c}^{(i)}) \land \textit{IL}_{q}(B_{d}^{(i)}))) \}\nonumber
\end{multline*}
where $\text{BX}(X)$ is the set of boxes over space $X$ and $\text{T}{=}\{(O,I),(I,O),(IO,IO)\}$.
See \cref{fig:reachabilityExample} for an
example drawn from analysis conducted on the model in \cref{subsubsec:model:cpu}. 
For feed-forward neural nets with non-decreasing activation functions, $B$ may be found by covering 
the input space, propagating boxes through the network, testing membership to $B$ of the 
resulting input- and output-boxes, and refining abstract states as needed over input-boxes that
produce output-boxes overlapping with $B$. Implementation details can be found in \cref{sec:networksAndAIDs}.
Note that we cover the relevant portions of the input space via iterative dissection informed by properties of the
problem, \textit{not} a na\:ive gridding of the
entire space unless repeated refinement has revealed that to be necessary.

The exact sizes of the boxes found by our \CEGAR-like process are determined by a series of hyper-parameters,
\footnote{For example, the maximum number of refinement iterations or the minimal size abstractions one is 
willing to consider;
for details on such hyper-parameters of \CEGAR\ and other bounded-model checking approaches, the interested reader may refer to
\cite{clarke2000counterexample,clarke2003verification,biere2003bounded}.} which the Fanoos maintains
in {\it states}, a fact we will return to in \cref{subsec:userFeedback}.

Prior to proceeding, the illuminated components of $B$ (i.e., the $\textit{IL}_{q}(B_{d}^{(i)})$ ) may undergo
some limited merging, particularly when an increase of abstraction level is sought.
Our merging process is closed over the family of abstract states we have selected; up to a numerical precision
threshold, boxes may only merge together to 
 form larger boxes, and only if the smaller boxes formed a partition of the larger box.
Value differences within the merging threshold are considered a match (i.e, a soft-match), and
 allow the pertinent sets of boxes to merge into 
larger boxes with slightly larger net volumes. Note that allowing abstract states to grow continues to make our
estimates conservative, and thus continues to ensure the soundness of Fanoos.
Absent any expansions due to soft-matching,
merging 
increases the size of abstract states  without anywhere increasing
the volume of their union  ---
this is not necessarily what would occur if one attempted the \CEGAR-like analysis again with
parameters promoting higher granularity. 
Essentially, merging here is one strategy of increasing abstraction level while retaining
some finer-resolution details.
As before, the \FanoosState maintains parameters 
which control the extent of this stage's merging. Optimal box-merging itself is an NP-hard task in general, so we
adopted roughly a greedy approximation scheme interlaced with hand-written heuristics for accelerating 
match-finding (e.g., feasibility checks via shared-vertex lookups) and parameters bounding the extent of computation.

\subsection{Generating Descriptions}
Having generated $B$, we produce an initial response, $r_0$, to the user's query in three steps as follows:
\begin{enumerate*}[{label=(\arabic*)}]  %
\item \label{gendescr-step1} for each member of $B$, we extract the box tuple 
members that were illuminated by $q$ (in the
case where $S_{IO}$ is illuminated, we produce a joint box over both tuple members), forming a 
set of joint boxes, $B'$; 
\item \label{gendescr-step2}  next, we heuristically search over predicates $P$ for members that describe box $B'$ and compute a set of predicates covering all boxes;
\item \label{gendescr-step3} finally, we format the box covering for user presentation. 
\end{enumerate*}
A sample result answer is shown in \cref{lst:initialanswer}, and details on steps \ref{gendescr-step2} and \ref{gendescr-step3} how to produce it follow below.
Pseudocode for the process can be found in \cref{subsec:generateDescription}.\footnote{
\label{footnote:methodology:pseudocodeMoreCorrectThanProseInCaseOfDiscrepency}
Generally speaking, in the event that any prose provided in this section
(\cref{sec:q3_methodology})  appears to be in discrepancy with
the pseudocode, the pseudocode should be taken as the more reliable / more likely accurate
source between the two.}

\begin{lstlisting}[
  caption={Initial answer to question in \cref{lst:idpquestion}. Also see \cref{subsec:FanoosAndPrettyUI} and \ref{sec:sup:extendedIO}.},
  label={lst:initialanswer},
	linerange={8-11},
	basicstyle=\footnotesize,
	breakatwhitespace=true,
	breaklines=true,
  postbreak=\mbox{{$\hookrightarrow$}\space}
]
(0.45789160, 0.61440409, 'x Near Normal Levels')
(0.31030792, 0.51991449, 'pole2Angle_rateOfChange Near Normal Levels')
(0.12008841, 0.37943400, 'pole1Angle_rateOfChange High')
(0.06128723, 0.22426058, 'pole2Angle Low')
\end{lstlisting}

\noindent 

\subsubsection{Producing a Covering of $B'$}
\label[subsubsection]{subsubsec:producingCoveringOfBPrime}

Our search over $P$ for members covering $B'$ is largely based around the greedy construction of a set covering using a carefully
designed candidate score. 

For each member $b \in B'$ we want to find a set of candidate predicates capable of describing the box and for which we would like to form a larger covering. 
We find a subset $P_b \subseteq P$ that is consistent with $b$ in that each member of $P_b$ passes the checks called for by $q_t$ when evaluated on $b$ (see the Description
column of \cref{tab:questionDescriptions}). 
This process is expedited by a feasibility check of each member of $P$ on a vector randomly sampled from $b$, prior to the expensive check for inclusion in $P_b$.
Having $P_b$, we filter the candidate set further to $P'_b$: members of $P_b$ that appear most specific 
to $b$; notice that in our setting, where predicates of varying abstraction level co-mingle in $P$, $P_b$ may contain many members that only loosely fit $b$. The subset $P'_b$ is formed by sampling outside of $b$
at increasing radii (in the $\ell_{\infty}$ sense), collecting those members of $P_b$ that fail to 
hold true at the earliest radius (see the pseudo-code in \cref{sec:appendix:psuedocode} for further details). 
Importantly, looking ahead to forming a full covering of $B$, if none of the predicates fail prior to exhausting\footnote{The operational meaning of ``exhausting'', as well as the radii sampled, are all parameters stored in the state.} this sampling, we report $P'_b$ as empty, allowing us to handle $b$ downstream as we will detail in a moment; 
this avoids having ``difficult'' boxes force the use of weak predicates that would ``wash out'' more granular details.
Generally speaking, we try to be specific at this phase under the assumption that the desired description granularity was determined earlier, 
presumably
during the \CEGAR-like analysis but also (in extensions to our approach) by altering\footnote{For example, filtering using a (possibly learned) taxonomy. 
Our current implementation has first-steps in this direction, allowing users to enable an optional operator that filters $P$
based on estimates of a  
model trained on previous interactions' data (see \cref{subsec:automatedPredicateSelectionDetails} for technical details). We comment 
further on this in \cref{subsec:userFeedback} and indicate why this operator is left as optional in \cref{sec:results}. 
 } $P$. For instance, if we want a subset of $P_b$ that was less specific to $b$ than $P'_b$, we might reperform the \CEGAR-like analysis so to produce larger abstract states.

To handle boxes for which $P'_b$ was empty, in general we insert into $P'_b$ a box-range predicate: a \emph{new} atomic
predicate that simply lists the variable ranges in the box (e.g., ``Box(x : [-1, 0], y: [0.5, 0.3])'').\footnote{Minimizing occurrences
of box-range predicates (if desirable) can be related to discussion in \cite{laird1986universal} regarding
the need for weak, universal rules to --- as the phrase goes --- prevent items failing through the cracks of
what is covered elsewhere.}%
As a result of providing cover for only one box, such predicates will only be retained by 
the (second) covering we perform in a moment
if no other predicates selected are capable of covering the box's axes. 
When a request to increase the abstraction level initially finds $P'_b$ empty, we may (as determined by \FanoosState parameters) set $P'_b$
equal to $P_b$ as opposed to introducing a box-range predicate.
If $P_b$ is empty as well, we are forced to add the novel predicate.
For reference later, we define $P''_b$ to be the singleton set containing the box range predicate added to $P'_b$ in the 
case $P'_b$ is extended, and the empty set otherwise (i.e., when $P'_b = P_b$).

We next leverage the $P'_b$ sets to construct a covering of $B'$, proceeding in an iterative greedy fashion.
Specifically, we form an {\it initial} covering  
\[K_f = \mathscr{C}_f(\cup_{b \in B'}\cup_{p \in P'_b}\{(p, b)\}, P)\]
 where   $\mathscr{C}_i(R, H)$
 is the covering established at iteration $i$, incrementing to 
\[\mathscr{C}_{i+1}(R, H) =  \mathscr{C}_{i}(R, H) \cup \Big\{ \text{argmax}_{p \in H \setminus \mathscr{C}_{i}(R, H)}\mu(p, \mathscr{C}_{i}(R, H), R)\Big\}\]
where $\mathscr{C}_{0}(R, H) = \emptyset$, $f$ is the iteration of convergence, and the cover score $\mu$ is
\[
\mu(p, \mathscr{C}_{i}(R, H), R)=\sum_{b \in B'}\mathbbm{1}(|\textsf{UV}(b, \mathscr{C}_{i}(R, H))\cap\textsf{FV}(p)| > 0)\mathbbm{1}(~(p,b) \in R)
\]
and $\mathsf{UV}(b,\mathscr{C}_{i}(R, H))$ is the set of variables in $b$ that are not constrained by $\mathscr{C}_{i}(R, H) \cap P_b$; since the boxes are multivariate and our predicates
typically constrain only a subset of the variables,
we select predicates based on how many boxes 
would have open variables covered by them. 
Pseudocode for this process can be found in algorithm \ref{algo:getApproximateMultivariateSetCoverPartA}.
Notice that $K_f$
is not necessarily an approximately minimal covering of $B$ with respect to members of 
$P$---by forcing $p \in P'_b$ when calculating the cover score $\mu$, we enforce additional 
specificity criteria that the covering should adhere to.
    At this stage, due to the nature of $P'_b$ being more specific than $P_b$, it is possible that
some members of $K_f$ cover one another --- that is, there may exist $p \in K_f$ such that $K_f \setminus \{p\}$ still
covers as much of $B'$ as $K_f$ did. By forming $K_f$, we have found a collection of predicates
that can cover $B'$ to the largest extent possible, selected based on how much of 
$B'$ they were \textit{specific} over (given by the first argument to $\mathscr{C}_f$ when forming $K_f$).
We now 
perform another covering, in order to
remove any predicates that
are dominated by the collection of other, potentially less-specific predicates that we had to include:
\[C_F = \mathscr{C}_F(\cup_{b \in B'}\cup_{p \in (P_b \cup P''_b)}\{(p, b)\}, K_f)\]
Here, similar to the subscript for $K_f$, the subscript $F$ is the iteration for which this covering converges.

\subsubsection{Cleaning and Formatting Output for User}

Having produced $C_F$, we collect the covering's content into a formula in DNF. 
If $b \in B'$ and $s$ is a maximal, non-singleton subset of $C_F \cap P_b$,
then we form a conjunction over the members of $s$, doing some filtering for conjunctions that would be implied by others.
Concretely, if $A = \cup_{b \in B'}\{ P_b \cap K'_F\}$, we construct:
\begin{align*} 
d_0 = \{ \land_{p \in s}p | (s \in A) \land  \lnot( ~ \exists s' \in A. s \subsetneq s'~)\} \enspace.
\end{align*}
Notice that the filtering done in $d_0$ is only a minor efficiency aid --- we do a final redundancy check momentarily
that, had we not filtered here, would have achieved similar results.
Ultimately, the members of $d_0$ are conjunctions of predicates,\footnote{From hereon, we use ``conjunct'' only to reference non-degenerate (i.e., not 1-ary or 0-ary) cases.} with their membership
to the set being a disjunction. Prior to actually converting $d_0$ to DNF,
we form $d'_0$ by: (1) removing any $c \in d_0$ that are redundant given the rest of
$d_0$ (see algorithm \ref{algo:removePredicatesImpliedByOthers})---in practice, $d_0$ is small enough to
simply do full one-vs-rest comparison and determine results with a SAT solver; (2) attempting to merge
any remaining box-range predicates into the minimal number necessary to cover the sets they are responsible for.
Note that this redundancy check is distinct from
our process to form $C_F$ out of $K_f$, since the latter worked at the abstract-state level (so it could not tell
if a disjunction of predicates covered a box when no individual predicate covered the entirety of the box) and attempted
to intelligently select predicates to retain based on maximizing a score.

Finally, $r_0$ is constructed by listing each $c$ that exists in $d'_0$ 
sorted by two relevance scores: first, the approximate proportion of the volume in $B'$
uniquely covered by $c$, and second by the approximate proportion of total volume $c$ covers in $B'$. 
The algorithm is shown in algorithm \ref{algo:getVolumesCoveredInformation}.
These sorting-scores can be thought of similarly to recall measures.
Specificity is more difficult to tackle, since it would
require determining the volume covered by each predicate (which may be an arbitrary
first-order formula) across the box bounding the universe, not just the hyper-cubes at hand;
this can be approximated for each predicate using set-inversion,
but requires non-trivial additional 
computation for each condition
(albeit work that may be amenable to one-time pre-computation up to a certain granularity). 

\subsection{User Feedback and Revaluation}
\label{subsec:userFeedback}

Based on the initial response $r_0$, users can request a more abstract or less abstract explanation. 
We view this alternate explanation generation as another heuristic search, where the
system searches over a series of \FanoosStates to find those that are deemed acceptable
by the user (consecutive user requests can be viewed in analogy to paths in a tree of Fanoos's \FanoosStates). 
The \FanoosStates primarily include algorithm hyper-parameters, the 
history of interaction,
 the question to be answered, and the set $B$. Abstraction and refinement operators take a current state and produce a new one, often
by adjusting the system hyper-parameters and recomputing $B$. This state-operator model 
of user response allows for rich styles of interaction with the user, beyond and
alongside of the three-valued responses of acceptance, increase, or decrease of the abstraction level shown in \cref{lst:lessabstract}.

\begin{lstlisting}[
  caption={Response to ``less abstract'' than \cref{lst:initialanswer}},
  label={lst:lessabstract},
	linerange={23-27},
	basicstyle=\footnotesize,
	breakatwhitespace=true,
	breaklines=true,
  postbreak=\mbox{{$\hookrightarrow$}\space}
]
(0.11771033, 0.12043966, 'And(pole1Angle Near Normal Levels, pole1Angle_rateOfChange Near Normal Levels, pole2Angle High, pole2Angle_rateOfChange Low, vx Low)')
(0.06948142, 0.07269412, 'And(pole1Angle High, pole1Angle_rateOfChange Near Normal Levels, pole2Angle_rateOfChange High, vx Low, x Near Normal Levels)')
(0.04513659, 0.06282974, 'And(endOfPole2_x Near Normal Levels, pole1Angle Low, pole1Angle_rateOfChange High, pole2Angle High, pole2Angle_rateOfChange Near Normal Levels, x High)')q
type letter followed by enter key: b - break and ask a different question,
 l - less abstract , m - more abstract, h - history travel
\end{lstlisting}

For instance, a history-travel operator allows the state (and
thus $r$) to return to an earlier point in the interaction process, if the user feels that
response was more informative; from there, the user may investigate
an alternate path of abstractions. Other implemented operators allow for refinements of 
specified parts of explanations
as opposed to the entire reply; the simplest form of this is by regenerating the explanation
without using a predicate that the user specified be ignored, while a more sophisticated operator
determines the predicates to filter out automatically by learning from past interaction. 
Underlying the discussion  of these mechanisms is the utilization of a concept of abstractness, a notion we further comment on in the next subsection.

As future work, we are exploring the use of active learning leveraging user interactions to select from a collection of operators, with particular interest in bootstrapping the learning process using
operationally defined oracles to approximate users. See \cref{subsec:sup:learningtoselectops} for further discussion.

\subsection{Capturing the Concept of Abstractness}
\label{subsec:methodologyCapturingConceptOfAbstractness}

The criteria to judge degree-of-abstractness in the lay sense are often difficult to capture.\footnote{See \cref{sec:appendix:pretheoreticalNotionsOfAbstractness}.}
We consider abstractness a diverse set of relations that subsume the part-of-whole
relation, and thus also generally includes the subset relation. 
For our purposes, defining this notion is not necessary, since we simply wish to 
utilize the fact of its existence. We understand abstractness to be a semantic
concept that shows itself by producing a partial ordering over semantic states (their ``abstractness''
level) which is in turn reflected in the lower-order semantics of the input-output
boxes, and ultimately is reflected in our syntax via explanations of different granularity.
Discussions of representative formalisms most relevant to computer science can be found in 
\cite{cousot1977abstract,10.1145/360303.360308,standardization1996iso,liskov1987keynote,liskov1994behavioral,DBLP:conf/uai/HostetlerFD15,DBLP:journals/cacm/Miller95}: \cite{cousot1977abstract} features abstraction in verification,
\cite{10.1145/360303.360308} features abstraction at play in
interpreting programs, \cite{standardization1996iso} is an excellent example of interfaces providing
 a notion of abstractness in network communications, \cite{liskov1987keynote,liskov1994behavioral} discuss notions
of abstractness relevant for type systems in object-oriented programming languages, 
\cite{DBLP:journals/cacm/Miller95} is a work at the intersection of NLP and knowledge 
representation that considers some pertinent concept relationships %
(hyponymy, meronymy, troponymy, etc.),\footnote{We point to \cite{DBLP:journals/cacm/Miller95} as 
a classic, widely utilized, and relatively approachable example from knowledge-representation-like endevours. In general,
modern works in knowledge engineering, ontologies, taxonomies, and so forth can be relevant here for similar reasons.}
and  \cite{DBLP:conf/uai/HostetlerFD15} shows an adaptive application in reinforcement learning.
An excellent discussion of the philosophical underpinnings and extensions can be found 
in \cite{floridi2008method}. 

In this work, the primary method of producing explanations at desired levels of abstraction 
is entirely implicit---that is, without explicitly tracking 
what boxes or predicates are considered more or less abstract (note that the operator
we mentioned that attempts to learn such relations is invoked optionally by human users, and it not 
used in any of the evaluations we present). 
Instead, we leverage the 
groundedness of our predicates to naturally form partial orderings over semantic states (their ``abstractness''
level) which in turn are appropriately reflected in syntax.

On the opposite end of the spectrum is explicit expert tuning    of abstraction orderings to be used in the system. 
Fanoos can easily be adapted to leverage labels from experts (e.g., taxonomies as in \cite{srikant1995mining}, or even simply type/grouping-labels without explicit hierarchical information) to preference 
subsets of predicates conditionally on user responses, but for the sake of this paper, we reserve agreement with human-provided labels as an independent metric of performance in our evaluation, prohibiting the free use of such knowledge by the algorithm during testing. 
As a side benefit, by forgoing direct supervision, we demonstrate that the concept of abstractness is recoverable from the semantics and structure of the problem itself.

\section{Related Work and Discussion}
\label{sec:q2_relatedWork}
Many methods are closely related to XAI, stemming from a diverse body of literature and 
various application domains, e.g., \cite{david1988design,KBSSurvey1995,agrawal1993mining,hayes2016autonomously,sreedharanplan,DBLP:conf/kdd/Ribeiro0G16,DBLP:conf/cav/KatzBDJK17,wellman2004theory,benz2005game}.
Numerous taxonomies of explanation families have been proposed~\cite{miller2018explanation,anjomshoae2019explainable,DBLP:conf/eccv/KimRDCA18,lipton2016mythos,KBSSurvey1995,adadi2018peeking,guidotti2019survey,biran2017explanation,friedrich2011taxonomy,richardson2018survey,ventocilla2018towards,chuang2012interpretation,roberts2018planning,papadimitriou2012generalized,hailesilassie2016rule,chakraborti2019explicability}, 
with popular divisions being (1) between
explanations that leverage internal mechanics of systems to generate descriptions 
(decompositional approaches) versus
those that exclusively leverage input-output relations (pedagogical)\footnote{We have also found a similar %
division referred to
as ''introspective'' explanations versus ``rationalizations'', such as in
\cite{DBLP:conf/eccv/KimRDCA18}.}, (2) the medium that comprises the explanation (such as with
most-predictive-features \cite{DBLP:conf/kdd/Ribeiro0G16}, 
summaries of internal states via finite-state-machines \cite{DBLP:conf/iclr/KoulFG19},
natural language descriptions \cite{hayes2016autonomously,DBLP:conf/eccv/KimRDCA18}
or even visual representations \cite{Huang2019,DBLP:conf/eccv/KimRDCA18}),
(3) theoretical criteria for a good explanation (see, for instance, \cite{miller2017explainable}),
 and (4) specificity and fidelity of explanation.
Of note, 
the vast majority of modern XAI methodologies developed for learned systems %
lack any formal guarantees regarding the correspondence between the 
explanations produced and the
learned component's true behavior (e.g., as pointed out by \cite{fern2020dontBeFooled}).

Rule-based systems such as expert systems,
 and work in the (high-level) planning community have a long history of producing 
explanations in various forms; notably, hierarchical planning \cite{hayes2016autonomously,mohseni2015interactive}
naturally lends itself to explanations of 
multiple abstraction levels. 
All these methods, however, canonically work on the symbolic level, 
making them inapplicable to most modern ML methods.
High fidelity, comprehensible rules describing data points can also be discovered
with weakly-consistent
inductive logic programming \cite{muggleton1999inductive} or association rule learning \cite{hipp2000algorithms,agrawal1993mining} typical 
in data-mining. However, these approaches are typically 
pedagogical---not designed to leverage access to the internals of the system---do not offer a variety of 
descriptions abstractions or strengths, and are typically not interactive.
While some extensions of association rule learning (e.g., 
\cite{srikant1995mining,han1999mining,han1995discovery}) do consider
multiple abstraction levels, they
 are still pedagogical and non-interactive. Further, they only describe subsets 
of the data they analyze \footnote{ While support and confidence thresholds may 
be set sufficiently low to ensure each transaction is described by at least one 
rule, the result would be a deluge of highly redundant, low-precision rules 
lacking most practical value (this may be considered the most extreme case of 
the ``rare itemset problem'' as discussed in \cite{liu1999mining}).} and only 
understand abstractness syntactically, requiring complete taxonomies be provided
 explicitly and up-front.
Our approach, by contrast, leverages semantic information, attempts to 
efficiently describe all relevant data instances, and produces descriptions that
are necessarily reflective of the mechanism under study.

Decision support systems~\cite{palaniappan2008clinical,Wasylewicz2019,eom2006survey,eom1998survey,eom1990survey} 
typically allow users to interactively
investigate data, with operations such as drill-ups in 
OLAP (OnLine Analytical Processing) cubes analogous to a simple form of 
abstraction in that setting. The 
typical notions of analysis, however, largely operate by truncating portions of
 data distributions and running analytics packages on selected subregions at 
user's requests, failing to leverage access to the data-generation mechanism 
when present, and failing to provide explicit abstractions or explicit guarantees about
 the material it presents.

More closely related to our work are approaches to formally
analyze neural networks to decompositionally extract rules, ensure safety, determine
decision stability, 
or otherwise certify / verify them,
which we discuss in more detail below. 
Techniques related to our inner-loop reachability analysis have been used for
stability or reachability analysis in systems that are otherwise hard to analyze analytically,
often in the interest of ensuring safety.
Reachability analysis for FFNNs
based on abstract interpretation domains, interval arithmetic, or set inversion has
been used in rule extraction and neural net stability analysis \cite{KBSSurvey1995,driescher1997checking,thrun1995extracting,wen1996neuralware}
and continues to be relevant, e.g.,
for verification of multi-layer perceptrons (MLPs) \cite{pulina2010abstraction,mirman2018differentiable,DBLP:conf/sp/GehrMDTCV18}, estimating the reachable states of closed-loop systems with MLPs in the loop \cite{DBLP:journals/corr/abs-1805-09944}, estimating the domain of validity of NNs \cite{DBLP:conf/ijcnn/AdamKMV15}, 
and analyzing security of NNs \cite{DBLP:conf/uss/WangPWYJ18}.
A similar variety of motivations and applications
exist for approaches
to NN verification and
rule extraction that are based on symbolic
decomposition of a network's units followed by constraint solving
or optimization  %
over the formulas extracted
\cite{taylor2005rule,etchells2006orthogonal,setiono1995understanding,towell1993extracting,murdoch2017automatic,tjeng2019evaluating,pmlr-v80-weng18a,10.1007/978-3-319-63387-9_1,DBLP:conf/uai/DvijothamSGMK18,DBLP:conf/cav/KatzBDJK17,DBLP:conf/atva/Ehlers17,DBLP:conf/nips/BunelTTKM18,DBLP:conf/nips/SinghGPV19}.
While these works provide methods to extract descriptions that faithfully
reflect behavior of the network, they do not generally consider
end-user comprehension of descriptions,
do not consider varying
description abstraction, and
do not explore the practice of strengthening
descriptions by ignoring the effects of measure-zero sets.
Also, many such techniques are only designed to characterize output behavior given particular input sets, whereas
we capture relations in multiple direction (i.e., input to output, output to input, and relations
between both simultaneously).

The high-level components of our approach can be compared to \cite{hayes2017improving}, where hand-tunable 
 rule-based methods with natural language interfaces encapsulate a module 
responsible for extracting information about the ML system, with explanation generation in part 
relying on minimal set-covering methods to find predicates capturing the model's states.
Extending this approach to generate more varying-resolution descriptions, however, does not seem
like a trivial endeavor, since (1) it is not clear that the system can appropriately handle 
predicates that are not logically independent, and expecting experts to explicitly know and 
encode all possible dependencies can be unrealistic, 
(2) the system described does not
have a method to vary the type of explanation provided for a given query when its 
initial response is unsatisfactory, and (3) the method produces explanations by first learning simpler models via Markov decision processes (MDPs).
Learning simpler models by sampling behavior
of more sophisticated models is an often-utilized, widely applicable method to bootstrap human understanding
(e.g. \cite{biran2017explanation,DBLP:conf/iclr/KoulFG19,gunning2017explainable}), but
it comes at the 
cost of failing to leverage substantial information from 
the internals of the targeted learned system. 
Crucially, such a technique cannot guarantee the fidelity of their explanations in respect to the learned system being explained, in contrast to our approach.

In \cite{perera2016dynamic},
 the authors develop vocabularies and circumstance-specific human models to 
determine the parameters of the desired levels of abstraction, specificity and location in robot-provided explanations about the robot's specific, previous experiences in terms of trajectories in a specific environment, as opposed to the more generally applicable conditional explanations about the internals of the learned component generated by Fanoos.
The particular notions of abstraction and granularity from multiple, distinct, unmixable vocabularies of \cite{perera2016dynamic} evaluate explanations in the context of their specific application and are not immediately applicable nor easily transferable to other domains.
Fanoos, by contrast, does not require  separate vocabularies and enables descriptions to include multiple abstraction levels (for example, mixing them as in the 
sentence ``House X and a 6m large patch on house Y both need to be painted'').

Closest in spirit to our work are the planning-related explanations of \cite{sreedharanplan},\footnote{We note that \cite{sreedharanplan} was published after the core of our approach was developed; both of our thinkings developed independently. 
\cref{sec:appendix:furtherMaterials} points to some relevant material.} providing multiple 
levels of abstraction with a user-in-the-loop refinement process,
but with a focus on markedly different search spaces, models of human interaction, algorithms
for description generation and extraction, and experiments. 
Further, we attempt to tackle the difficult problem of extracting high-level symbolic knowledge 
from systems where such concepts are not natively embedded, in contrast to \cite{sreedharanplan},
who consider purely symbolic systems. 

In summary, current approaches focus on single aspects of explanations, fixed levels of abstraction, or provide inflexible guarantees (if any) about the explanations given. 
With Fanoos, we demonstrate the feasibility of achieving
the desired flexibility in explanations and %
adjustability of fidelity level
using methods that interleave between formal verification techniques, search 
heuristics, and user interaction. %

\section{Experiments and Results}
\label{sec:results}
In this section we discuss empirical demonstrations of Fanoos's ability to produce and adjust descriptions across two different domains. 
The code implementing our method and supporting information 
(e.g., predicate definitions) can be found in our code repo, \url{https://github.com/DBay-ani/Fanoos}.
Our presentation keeps with norms presently established in XAI work presented across multiple venues (e.g., \cite{aha2017ijcai} and section 4.8 of \cite{anjomshoae2019explainable}). 

\subsection{Systems Analyzed}

We analyze learned systems from robotics control and
more traditional ML predictors to demonstrate the applicability to diverse domains.
Information on the predicates available for each domain
can be found in \cref{tab:predsPerDomain}.

\begin{table}
\caption{Summary statistics of predicates in each domain.
For description of MA/LA labels, see \cref{subsec:expertLabeling}.
Percentages are rounded to three decimal places.
    }
\label{tab:predsPerDomain}
\begin{tabularx}{\columnwidth}{Xcc}
\toprule
    & CPU  & IDP
\\
\midrule
    Input space predicates & 33 & 62
\\
    Output space predicates & 19 & 12
\\
    Joint input-output space predicates & 8 & 4
\\
    Total with MA (more abstract) label & 20 & 15
\\
    Total with LA (less abstract) label & 40 & 63
\\
      Percentage of Pred.s Labeled MA & 0.333 & 0.192 
\\
\bottomrule
\end{tabularx}
\end{table}

\subsubsection{Inverted Double Pendulum (IDP)}
\label[subsubsection]{subsubsec:model:IDP}
The control policy for an inverted double-pendulum is tasked to keep a pole steady and upright; the pole consists of two under-actuated segments attached end-to-end, rotationally free in the same plane; the only actuated component is a cart with the pivot point of the lower segment attached. 
While similar to the basic inverted single pendulum example in control,
this setting is substantially more complicated, since multi-pendulum systems are known to exhibit chaotic behavior \cite{kellert1993wake,levien1993double}. The trained policy was taken from reinforcement learning literature \cite{schulman2017proximal,rl-zoo}.\footnote{The policy was trained using PPO2~\cite{schulman2017proximal}
 which, as an actor-critic method, uses one network to produce the action, and one to estimate state-action values.}.
The seven-dimensional observation space 
--- the bounding box for which can be found in \cref{tab:invertedDoublePendulumBB} located in \cref{sec:appendix:boundingBoxes} ---
includes the segment's angles, the cart x-position, their time derivatives, and the y-coordinate of the second pole.
The output is a torque in $[-1, 1]\text{Nm}$ and a state-value estimate, which is not 
a priori bounded.
Internal to the analyzed model is a transformation to
convert the observations we provide to the form expected by the networks---chiefly, 
the angles are converted to sines and cosines and observations are standardized in 
respect to the mean and standard deviation of the model's training runs.
The values chosen for the input space bounding box were inspired by the 5\% and 
95\% quantile values over simulated runs of the model in the \textsf{rl-zoo} framework. 
We expanded the input box beyond this range to allow for the examination of rare inputs and 
observations the model was not necessarily trained on;\footnote{For instance, the implemented train and test environments exit whenever the end of the second segment is below a certain height.
In real applications, a user may 
wish to check that  
a sensible recovery is attempted
after entering such unseen situations.} whether
the analysis stays in the region trained-over depends on the user's question.

\subsubsection{CPU Usage (CPU)}
\label[subsubsection]{subsubsec:model:cpu}
We also analyze a more traditional ML algorithm for a non-control task --- a polynomial regression for modeling CPU usage. 
Specifically, we use a three-degree fully polynomial basis 
over a 5-dimensional input space\footnote{The input space includes cross-terms and
the zero-degree element---e.g., $x^2y$ and $1$ are members.} to linearly regress-out a 
three-dimensional vector.
We trained our model using the publicly available data from \cite{OpenML2013}\footnote{
Dataset available at \url{https://www.openml.org/api/v1/json/data/562}}.
The observations are 
[\textit{lread}, \textit{scall}, \textit{sread}, \textit{freemem}, \textit{freeswap}] and the response variables we predict are
$[\textit{lwrite}, \textit{swrite}, \textit{usr}]$.
For analysis convenience, we normalized the input space in respect to the
training set min and max prior to featurization, which was the same method of 
normalization used during model training and evaluation;
alternatively, Fanoos could have used the raw (unnormalized) feature space and insert the normalization as part of the model-pipeline. 
We opted to analyze an algorithm with a degree-3 polynomial feature-set after normalizing
the data in respect to the minimum and maximum of the training set
since this achieved the highest performance---over 90\% accuracy---on a 90\%-10\% train-test split of the data compared to similar models with 1,2, or 4 degree
features\footnote{Note that while we did do due-diligence in producing a performant and
soundly-trained model, the primary point is to produce a model worthy of analysis.}.
While the weights of the regression may be interpreted in some sense (such as indicating which
individual feature is, by itself, most influential), the
joint correlation between the features and non-linear transformations of the input values makes it far from clear how the model behaves over the original input space.
For Fanoos, the input space bounding box was determined from the 5\% and 95\% quantiles for each input-variable over the full, normalized dataset; the exact values can be found in \cref{tab:CPUUsageBB} located in \cref{sec:appendix:boundingBoxes}.

\subsection{Experiment Design}
\label{subsec:experimentDesign}
Tests were conducted using synthetically generated 
user interactions, with the goal of determining whether our approach properly
changes the description abstractness in response to the user request. 
The domain and question type were randomly chosen, the latter selected among the options listed in \cref{tab:questionDescriptions}.
The questions themselves were randomly generated to have up to four disjuncts, each with
conjuncts of length no more than four; conjuncts were ensured to be distinct, and
only predicates respecting the constraints of the question type were used.
After posing an initial question, interaction with Fanoos  was randomly selected from 
four alternatives (here, MA means ``more abstract'' and LA means ``less abstract''):
Initial refinement of $0.25$ or $0.20$ $\rightarrow$ make LA $\rightarrow$ make MA $\rightarrow$ exit;
Initial refinement of $0.125$ or $0.10$ $\rightarrow$ make MA $\rightarrow$ make LA $\rightarrow$ exit.
For the results presented here, over 130 interactions were held, 
resulting in several hundred responses from Fanoos.

\subsection{Metrics}
\label{subsec:metrics}
We evaluated the abstractness of each
response from Fanoos using metrics across the following categories: reachability analysis, 
description structure, and human word labeling.

\subsubsection{Reachability Analysis}
We compare the reachability analysis results
when producing descriptions of different abstraction levels, 
which call for different levels of refinement. Specifically,
we record statistics about the input-boxes generated
during the \CEGAR-like analysis after normalizing them in respect to the input space bounding box so that each axis is in $[0,1]$,
yielding results which are comparable across domains. The following values are examined:
\begin{itemize}
\item Volume of the box (product of its side lengths).
\item Sum of the box side lengths. Unlike the box volume, this measure is at least as large as the maximum side length. 
\item Number of boxes used to form the description.
\end{itemize}
The volume and summed-side-lengths are distributions, reported in terms of
the summary statistics shown in \cref{tab:mainResults}.
These metrics provide a rough sense of the abstractness notion implicit in the size of boxes and how they relate to descriptions.

\subsubsection{Description Structure}
Fanoos responds to users with a multi-weighted DNF description. 
This structure is summarized as follows to give a rough sense of how specific each description is by itself:
\begin{itemize}
\item Number of disjuncts, including atomic predicates
\item Number of non-singleton conjuncts, providing a rough measure of the number of ``complex'' terms
\item Number of distinct 
    named predicates: atomic, user-defined
    predicates that occur anywhere in the description (even in conjunctions).
    This excludes box-range predicates.
\item Number of box-range predicates that occur anywhere (i.e., in conjuncts as well as stand-alone).
\end{itemize}

The Jaccard score and overlap coefficients below are used to measure similarity in the verbiage used in two descriptions.

\begin{itemize}
\item Jaccard score: general similarity between two descriptions, viewing the set of atomic predicates used in each description as a bag-of-words.
\item Overlap coefficient: measures whether
    one description is simply a more ``verbose'' variant of the other, in the sense
    that the set of atomic predicates of one is a subset of the other using $\frac{| S_1 \cap S_2|}{min(|S_1|, |S_2|)}$, 
    where $S_1$ and $S_2$ are the sets of predicates used in the two descriptions.
\end{itemize}

\subsubsection{Human Word Labeling}
\label[subsubsection]{subsec:expertLabeling}

As humans, we grasp  which atomic
predicates map to comparatively more abstract notions in the world,
and as such can evaluate the responses based on usage of more vs. less abstract verbiage. 
It is important to note that this approach---on descriptions built from atomic predicates---yields an informative approximation rather than a true measure of abstractness for the following reasons: it is not 
clear that the abstractness of a description's components translates in an obvious fashion to the abstractness of the whole (in a similar vein, we do not rule out the possibility that predicates of the same level in the partially ordered set of abstractness can be combined to descriptions of different abstractness\footnote{For example, just because
two description use verbiage from the same human-labeled category of abstractness, 
it does not mean the two descriptions have the same level of abstractness.}).
This phenomenon becomes more pronounced in coarsely grained partitions, where nuances are hidden in the partitions. 
For simplicity we choose two classes, more abstract (MA) vs. less abstract (LA), in the measures below: 
\begin{itemize}
\item Number of predicates accounting for multiplicity, i.e., if an atomic
    predicate $q$ has label MA and occurs twice in a sentence, it contributes
    two to this score.
\item Number of unique predicates: e.g., if an atomic
    predicate $q$ has label MA and occurs twice in a sentence, it contributes
    one to this score.
\item Prevalence: ratio of unique predicates to the 
    total number of atomic predicates in a description. This measure is particularly 
    useful when aggregating the results of multiple descriptions into one distribution, 
    since the occurrence of predicates is statistically coupled with the length of descriptions;
    under a null hypothesis of random generation of content, one would expect longer sentences
    to contain more MA,LA predicates, but expect the proportion to remain constant.
\end{itemize}
Each of the above measures have obvious counter-parts for predicates with MA/LA labels. We
note that prevalence will not necessarily sum to $1$, since box-range predicates are atomic predicates without either label.

\subsection{Results}
Running the experiments described in \cref{subsec:experimentDesign},
we collected a series of states and the summary statistics on them 
described in \cref{subsec:metrics}. Since we are chiefly interested in 
evaluating whether a description changes to reflect the abstraction
requested by the user, we examine the relative change in response to user interaction. 
Specifically, for pre-interaction state $S_t$ and post-interaction state $S_{t+1}$, we collect metrics $m(S_{t+1}) - m(S_t)$, describing \emph{relative} change 
for each domain-response combination. 
This same process is used for the  Jaccard score and overlap coefficients, except the values in question are computed as $m(S_{t+1}, S_t)$.
The medians of these distributions are reported in \cref{tab:mainResults}.

\begin{table}[t]
\caption{Median relative change in description before and after Fanoos adjusts the abstraction in the requested direction
Results are rounded to three decimal places. Further notes in \cref{sec:sup:extendedIO}. }
\label{tab:mainResults}
\begin{tabularx}{\columnwidth}{ccX@{\hspace{0em}}r@{\hspace{0.5em}}r@{\hspace{0.5em}}r@{\hspace{0.5em}}r}
\toprule
& & & CPU & CPU & IDP & IDP
\\ \cline{3-7}
&& Request & LA & MA & LA & MA
\\ \hline
\multirow{5}{*}{\rotatebox[origin=c]{90}{Reachability}} & Boxes & Number & 8417.5 & -8678.0 & 2.0 & -16.0
\\ \cline{2-7}
& \multirow{4}{*}{\rotatebox[origin=c]{90}{\shortstack{Sum side\\lengths}}} & 
Max & -1.125 & 1.125 & -1.625 & 1.625
\\
& & 
Median & -1.187 & 1.188 & -2.451 & 2.438
\\
& & 
Min & -0.979 & 0.986 & -2.556 & 2.556
\\
& & 
Sum & 21668.865 & -22131.937 & 582.549 & -553.007
\\ \cline{2-7}
& \multirow{4}{*}{\rotatebox[origin=c]{90}{Volume}} & 
Max & -0.015 & 0.015 & -0.004 & 0.004
\\
& & 
Median & -0.003 & 0.003 & -0.004 & 0.004
\\
& & 
Min & -0.001 & 0.001 & -0.003 & 0.003
\\
& & 
Sum & -0.03 & 0.03 & -0.168 & 0.166
\\ \midrule
\multirow{6}{*}{\rotatebox[origin=c]{90}{Structural}} & \multicolumn{2}{l}{Jaccard} & 0.106 & 0.211 & 0.056 & 0.056
\\ 
& \multicolumn{2}{l}{Overlap coeff.} & 0.5 & 0.714 & 0.25 & 0.25
\\
& \multicolumn{2}{l}{Non-Singleton Conjuncts} & 1.0 & -2.0 & 0.5 & -2.5
\\ 
& \multicolumn{2}{l}{Disjuncts} & 7.0 & -7.5 & 2.0 & -2.5
\\ 
& \multicolumn{2}{l}{Named preds.} & 1.0 & -1.0 & 1.0 & -4.5
\\
& \multicolumn{2}{l}{Box-Range preds.} & 2.0 & -2.0 & 1.5 & -1.5
\\ \midrule
\multirow{5}{*}{\rotatebox[origin=c]{90}{Words}} &
\multirow{3}{*}{\rotatebox[origin=c]{90}{\shortstack{MA\\terms}}} 
 & Multiplicity & 3.0 & -3.0 & 24.0 & -20.0
\\
& & Uniqueness & 0.0 & 0.0 & 1.0 & -1.5
\\
& & Prevalence & -0.018 & 0.014 & -0.75 & 0.771
\\ \cline{2-7}
& \multirow{3}{*}{\rotatebox[origin=c]{90}{\shortstack{LA\\terms}}} 
& Multiplicity & 20.0 & -21.5 & 68.5 & -86.0
\\
& & Uniqueness & 2.0 & -2.0 & 12.0 & -14.0
\\
& & Prevalence & 0.0 & 0.0 & 0.0 & 0.0
\\ \bottomrule
\end{tabularx}
\end{table}

In summary, the reachability and structural metrics follow the desired trends:
when the user requests greater abstraction (MA), the boxes become larger, 
and the sentences become structurally less complex---namely, they become shorter (fewer 
disjuncts), have disjuncts that are less complicated (fewer explicit conjuncts, hence
more atomic predicates), use fewer
unique terms overall (reduction in named predicates) and resort less often to referring to the
exact values of a box (reduction in box-range predicates). Symmetric statements can be made for when requests for less abstraction (LA)
are issued. From the overlap and Jaccard scores, we can see that the changes in response complexity
are not simply due to increased verbosity---simply adding or removing phrases to the descriptions from the prior steps---but also the result of changes in the verbiage used; this is appropriate since
abstractness is not exclusively a function of description specificity. 

    Trends for the human word-labels are similar, though more subtle to interpret. 
We see that use of LA-related terms follows the trend of user requests with respect to multiplicity and uniqueness counts (increases for LA-requests, decreases for MA-requests), while being less clear with respect to prevalence (uniform 0 scores). 
For use of MA terms, we see that the prevalence is correlated with user requests in the expected fashion (decrease on LA requests, increase on MA requests). 
Further, we see that this correlation is mirrored for the MA counts when taken relative to the same measures for LA terms.
Specifically, when a user requests greater abstraction (MA), the counts for LA terms decrease far more than those of MA terms, and the symmetric situation occurs for requests of lower abstraction (LA), as expected. 
While they depict encouraging trends, we take these human word-label measures with caution, due to the fragility of reasoning about the complete description's abstractness based on its constituents (recall that the abstractness of a description is not necessarily directly linked to the abstractness of its components).
Nevertheless, these results---labelings coupled with the structural trends---lend solid support that Fanoos can recover elements of an human's notion about abstractness by leveraging the grounded semantics of the predicates.

\section{Conclusions And Future Work}
\label{sec:conclusion}
Fanoos is an explanatory framework for ML systems that mixes technologies
ranging from classical verification to machine learning. %
Our experiments support that Fanoos can produce and navigate explanations at multiple granularities and strengths.
We are investigating operator-selection learning
and further data-driven predicate generation to accelerate knowledge base construction --- the latter focusing on representational power,
extrapolation intuitiveness, behavioral certainty, and data efficiency.
Finally, this work can adopt engineering
improvements to ML-specific reachability computations.
Further discussion is in \cref{sec:sup:fanoosextensions}.
We will continue to explore Fanoos's potential, and hope that the community finds inspiration in both the methodology and philosophical underpinnings presented here.

\bigskip
\noindent\textbf{Acknowledgments.}
We thank: Nicholay Topin for supporting our spirits at some key junctures of this work;
David Held for pointing us to the rl-baselines-zoo repository;
David Eckhardt for his proof-reading of earlier versions of this document; 
the anonymous reviewers for their thoughtful feedback.

\bibliographystyle{splncs04}
\bibliography{Bibliography-File}

\clearpage

\appendix

\section{Fanoos Extensions}
\label{sec:sup:fanoosextensions}

{\it Note:} This section of the appendix, \cref{sec:sup:fanoosextensions}, first appeared in \cite{IJCAIXAIFanoos2020},
published electronically on September $2^{\text{nd}}$, 2020 ;
we provide it here, with no note-worthy modifications, for the sake of completeness and record on ArXiv. 

{\it Update \dateUpdatePostVMCAITwentyTwentyTwo:} Some additional content has been added to this section. Much of the material added had been intended for inclusion in the VMCAI 2022 publication (\cite{DBLP:conf/vmcai/BayaniM22}), but was dropped due to space constraints. To see which content is new, we suggest comparing the Latex files present on ArXiv before and after this update using a file-diff utility.\\ \\

Here, we elaborate further on-going work for extensions of Fanoos in three thrusts:
operator-selection learning (\ref{subsec:sup:learningtoselectops}),
more advanced data-driven predicate generation (\ref{subsec:sup:predicategen}), and 
engineering improvements (\ref{subsec:supplementalreachability}, \ref{subsec:moreAboutNotionOfUssuallyTrue}, and \ref{subsec:FanoosAndPrettyUI}).

\subsection{Learning to Select Operators}
\label{subsec:sup:learningtoselectops}
Fanoos lends itself to the use of active learning to 
improve the operator-selection procedure, utilizing a proxy-user to help bootstrap the process.
Initial developments include an approach that attempts to 
balance exploration and exploitation, use a priori knowledge of the process and potentially the domain,
and leverage structure of the states.\footnote{While these 
three desiderata have tangled interrelations, they do not seem necessarily redundant: the first two seem possible without the third
in a problem possessing
bandit-esque labeled levers and an opaque oracle, while the first and third seem able to exclude the second if one simply
has rewards and distance measures between states.}
We are in the process of carrying out and refining experiments; we 
hope to report on this work in the near future.

\subsection{Predicate Generation}
\label{subsec:sup:predicategen}

We view the predicate generation problem as important, but distinct from the challenge
Fanoos attempts to tackle and one that is amenable to many solutions in context (
for instance, the methods we adopted in \cref{subsubsec:model:IDP} backed by elementary statistics).
Importantly, we do not believe
requiring predicates simply pushes our 
core problem
to a different arena. Our belief is that Fanoos addresses
a large swath of the desiderata and 
resulting 
pipeline 
while allowing 
improvements to be plugged in for specific subproblems - subproblems with reasonably mature literature, existing methods, and 
communities 
working there-in
to provide enhancements. Grounded predicate generation is one such subproblem. 
To elaborate with illustrative examples:
databases are considered sensible solutions
despite the fact that query optimizers can often be improved, while it is not sensible to say that the
traveling salesman problem has been solved 
on the basis of it reducing to the knapsack problem;
we view Fanoos's relationship to the predicate generation problem as similar to the former, not the latter.
That noted, we ultimately want an entire system that produces
well-tailored explanations for ML users
while requiring as little human effort as possible - particularly
effort that is unintuitive or requires uncommon expertise.
Achieving this greater ease of deployment in more circumstances requires making the subproblems -
such as grounded predicate generation - more effortless, more often.

On this front, we are further investigating learning techniques for the generation of domain predicates.
Among these, Generalized Hough Transforms~\cite{ballard1981generalising} were an early candidate;
worthy of note in relation to this are methods for automatic templates 
production (e.g., \cite{Lee00generalizedhough}).
More recently our attention has been drawn to 
advancements in inductive logic programming \cite{muggleton1999inductive}, Metagol~\cite{muggleton2014meta,metagol} chief among them.
These methods have the potential to provide representational flexibility, predicates amenable to human review, intuitiveness of the extrapolations that may be  necessary in the process of generating predicates, and data-efficiency.
With Metagol there also is potential for incorporating human-desirable invariants into predicates automatically, and handle input data with a spectrum of structure; in our experiments thus far, however, non-trivial 
effort has been required to handle numerical values as desired.
That noted, there exist extensions of Metagol and various algorithms in related logic paradigms which claim to be capable of handling (and scaling to)
image, spatial, and temporal data ( for example,
\cite{DBLP:journals/cacm/GulwaniHKMSZ15,DBLP:conf/lpnmr/WalegaBS15,DBLP:conf/ilp/SuchanBS16,10.1007/978-3-030-43823-4_16,DBLP:conf/ilp/DaiMWTZ17,DBLP:journals/ml/MuggletonDSTWZ18,DBLP:conf/ilp/DaiMZ15,srinivasan2001aleph,Suchan2020,DBLP:conf/cosit/BhattLS11}).
Further comments are made in \cref{subsubsec:predicateGeneration:furtherComments} below.

It is particularly desirable to enable the generation of predicates that are
invariant under certain types of user-defined transformations. 
For example, one might
learn the concept of closing a left-gripper and generalize it to closing at least one
gripper by
making the truth of the predicate invariant to renaming of hands. 
We might learn
the concept of a sharp turn by only observing left turns and ensure that the
hypothesis produced did not become curtailed to leftness by enabling invariance under rotation
and reflection (possibly as a wrapper function that converts turns, using these transforms, to left turns).
To keep such an approach both practical and meaningful, we must restraint its scope to not venture into more general AI challenges,
such as attempting to solve the whole of transfer learning.
Implementing invariance relating to naming, copying, and basic geometric transformations seems naturally desirable and
not unmanageable in common use cases, and is thus one of the sub-focuses of our research.

The space of possible solutions in the modern ML landscape to engineer predicate generation is vast, 
particularly if one is willing to admit predicates that work reliably in practice but are not necessarily perfect in all circumstances.\footnote{
This might involve, for instance, relaxing one's confidence in a predicate's reliability and/or asking Fanoos questions about usual behavior instead of worst-case behavior.}
For instance, one might utilize an oft-deployed vision-based recognition technique
 to pick-out objects reliably, despite it not being adversarially robust;
indeed, this is not
dissimilar to the approach adopted by \cite{DBLP:conf/eccv/KimRDCA18}, for example.
There might not be a panacea to the predicate generation challenge when it is conceived too broadly,
especially if one insists on retaining certainty, transparency, and data-efficiency while attempting to 
capture sophisticated concepts with minimal human effort.
Even in the face of this, we maintain that it is reasonable to suppose that understanding 
of a target system can most often (in practice) be improved through
analysis/ decomposition utilizing these potentially imperfect materials.\footnote{
It may be necessary to conduct the analysis recursively on components.}

Given this overview of how predicate generation may be supported,
it is worth taking a moment to reflect on whether predicates of this style are fundamentally
necessary to the project of ML explainability.
Not all XAI systems may require {\it explicit} semantic-grounding mechanisms that are {\it separate} from the learned system being analyzed---for instance, 
ranking features by weight,\footnote{Whether this counts as an explanation, particularly
if the features fail to have clear meaning to the consumer, is debatable.
Further, one could rank the weights of the polynomial regression we experimented on, but that would fail to address
all of the difficulties we highlighted in \cref{subsubsec:model:cpu}.
}
generic salience maps, LIME \cite{DBLP:conf/kdd/Ribeiro0G16},  and finding exemplar datum \cite{Huang2019}
 do not seem to utilize such mechanisms under casual examination.
Further, the non-necessity of these elements 
seems intuitively true\footnote{We note that ``intuitive'' here is not cause for added certainty, since people's intuitions about the organization of cognition, etc., are not always spot-on.} if we momentarily indulge in considering how a person explains their
behavior---though of course people produce explanations of questionable truthfulness, reliability, and accuracy when it comes
to both daily activity and deeper psychological phenomena.
 Necessity aside, 
explicit grounding approaches in the vein of this subsection are not uncommon in XAI efforts,
and for the foreseeable future
will most likely have fundamental benefits and drawbacks compared to the alternative.

\subsubsection{Further Comments On Methods for Predicate Learning, \dateUpdatePostVMCAITwentyTwentyTwo\\}
\label[subsubsection]{subsubsec:predicateGeneration:furtherComments}

It is natural to consider --- almost to the point of being unnecessary to explicitly mention --- what role ``more traditional and mainstream''
rule-producing ML techniques can play in the predicate generation problem.
Learning algorithms that are relatively data efficient and are typically
considered interpretable can apply, such as
decision trees (not only  standard Quinlan trees \cite{DBLP:books/mk/Quinlan93}, but also the
notable improvements over the last several decades, such as \cite{DBLP:conf/nips/HuRS19})  and 
advances drawing on techniques from
association rule learning, such as Graphs of Temporal Constraint (GTCs) \cite{DBLP:journals/jmlr/Guillame-BertD17}
which can efficiently learn often small directed graphs with attributed nodes and edges for classification of multi-channel temporal data instances. These techniques
\textit{can} differ sufficiently from ILP methods in some crucial ways --- for instance, in respect to data efficiency claims, strength of consistency with
training data, incorporation of background knowledge, expressive richness, and (potential) decreases in interpretability at scale --- that
grouping 
them into a separate profile %
makes sense.\footnote{At the least, %
in some qualitative sense,
the total variance of the ML methods 
mentioned in this paragraph over such criteria axes is generally larger --- which further supports considering them as at least one separate group. Elaborating on this ``variance'' with an example: whether or not one prunes a decision tree can introduce some ``spread'' in respect to a few of the 
listed criteria --- such as strong consistency with training data.  }
Worth noting, however, is that
these methods produce logical formulas of a form that tend to be both ingestable and tractable with relative ease for modern SAT solvers.
In a sense, we overall consider the methods mentioned in this subsection (subsection \ref{subsubsec:predicateGeneration:furtherComments}) ``somewhere between'' the 
aforementioned ILP-etc. methods
and the
``not perfect by often reliable in practice'' approaches %
in respect
to their profile of benefits and drawbacks for this application.\footnote{Quotation marks are here meant to suggest a 
spiritual, intuitive, or metaphorical sense as opposed to any sort of commitment to a claim situated in a  rigorous formality.}

\subsection{Reachability Analysis Performance}
\label{subsec:supplementalreachability}
The reachability analyzer of Fanoos is designed in a generic fashion and 
amenable to having its implementation swapped out without fundamentally altering the overall approach.
While reachability analysis is in principle computationally expensive, there are many algorithms that have undesirable worse-case bounds in theory---for example SAT solvers and the simplex algorithm---but routinely demonstrate useful performance in practice.
Methods to potentially draw upon for engineering improvements include the reachability toolboxes CORA \cite{DBLP:conf/cpsweek/Althoff15} and SpaceEx \cite{frehse2011spaceex}, as well as FaSTrack \cite{herbert2017fastrack,fridovich2018planning}, a safe planning framework that addresses a related family of problems; all have pushed forward the frontier of practical applications.
Naturally, %
in addition to these well-developed packages, %
one can build off of the 
assortment of academic literature --- such as \cite{DBLP:conf/sp/GehrMDTCV18} and others mentioned in \cref{sec:q2_relatedWork} ---
that improve the efficiency of
reachability techniques pertinent to systems of our interest.

\subsection{The Sampling Process Backing ``Usually True'' Queries}
\label{subsec:moreAboutNotionOfUssuallyTrue}

Our sampling method can be modified to user-provided
distributions, such as estimates of the typical input distribution.
 Currently, we use a uniform distribution over the hypercubes
where the condition is potentially true;  this can help examine the range
of behaviors/circumstances to a fuller extent
 and enable counter-factual reasoning.
We can envision both sampling approaches as providing distinct values in practice, and thus both potentially useful over the course of dealing with a learned system.

Note that areas of the input space that are logically impossible and have non-zero measure
under the uniform measure can be ignored by instructing Fanoos to add a predicate to ``what do you usually do when \ldots'' queries and subsequently filtering by that predicate
when forming boxes in response to ``when do you usually \ldots'' queries.\footnote{Both of these operations
could be supported in a push-button fashion under the hood of a UI.}
In the output space, Fanoos attempts to deal with the 
pushforward of the provided learned component under the input distribution. It is worth noting, then, that for some applications, Fanoos in a sense characterizes what the learned component ``attempts to do''
or ``signals'' as opposed to what occurs down-stream, which would not atypically involve additional contributors
to the system outside of Fanoos's purview\footnote{We do not rule-out a user developing a model for down-stream behavior and incorporating it
into something visible to Fanoos---arguably our experiments show signs of that by having actuator limits
for the inverted double-pendulum---but naturally this is not something to assume is always present in sufficient fidelity.}. For instance, Fanoos may be able to say that the controller of an
 autonomous vehicle ``attempts a hard-stop'',
but it may be the case that the vehicle as a whole exhibits a non-identical behavior due to ice on the road,
unmodeled performance of low-level controllers, or other conditions either internal or external to the machine.  

\subsection{Fanoos's User Interface}
\label{subsec:FanoosAndPrettyUI}

Our focus while developing Fanoos has been to ensure that the desired information 
can be generated.
In application, a thin front-end can be developed
to provide a more aesthetically pleasing presentation, using the vast array of infographics and related tools available (e.g., \cite{DBLP:conf/vissym/SharmaSPAS14}).
Various input-output wrapper packages from the extensive literature on decision support systems may provide useful guidelines,
as well as particularly promising works in interactive robotics (for instance, \cite{mohseni2015interactive}). 
Presenting results
as English-like sentences using templates (similar to \cite{hayes2017improving}),
as bar-graphs sorted on height, or
word-clouds emphasizing relative importance
 are all easily facilitated from Fanoos's output.
For input, 
template-supported English-like phrases, drag-and-drop flowcharts,
\footnote{A particularly user-friendly and open-source example being \cite{10.1145/1868358.1868363}.}
or even basic HTML drop-down menus with sub-categories for option filtering\footnote{Either grouped by variables of interest or some
richer semantic notion.}
are all possibilities.
Although certainly important in practical application, in this paper we rather focus on presenting the underlying algorithmic contributions than the user-facing presentation.

\section{Fanoos Structural Overview}
\label{sec:appendix:diagram}
\begin{figure}[h!]
\includegraphics[width=(0.8\linewidth)]{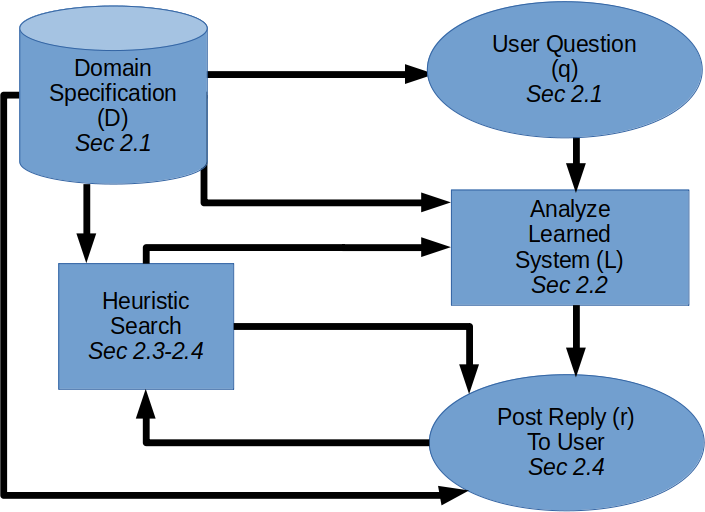}
\centering
        \caption{User-interaction with Fanoos}
        \label{fig:approachflowchart}
\end{figure}

\cref{fig:approachflowchart} illustrates the component interactions in
Fanoos. Sections detailing the component are italicized. Components
requiring user interaction are oval, internal modules are rectangular,
and the knowledge database cylindrical.

\section{Extended Example User Interactions}
\label{sec:sup:extendedIO}

We present here a typical user interaction with our system, \cref{fig:userioidp}. 
The interested reader can find
the predicate definitions with the code at both \url{https://github.com/DBay-ani/Fanoos} and \cite{david_bayani_2021_5513079}.
In practice, if users want to know more about the operational meaning of predicates (e.g., the exact conditions that 
each predicate tests for), open-on-click hyperlinks and/or hover text displaying the 
relevant information from the domain definition\footnote{Or derived from the
domain information and simplified/summarized.} can be added to the UI

Notice that our code uses a Unix-style interaction in the spirit of the \texttt{more} command, so not to flood the screen beyond
a preset line limit. We also support auto-complete, listing options available in the context and finishing tokens when unambiguous whenever the user hits tab.\footnote{For instance, suggestions / completions for predicates obey restrictions imposed by  \cref{tab:questionDescriptions} based on the question type thus-far specified by the user. }
In examples shown, whenever we insert a comment in the interface trace that was not originally there, we put
\texttt{//} at the beginning of the line.

In \cref{fig:userioidp}, we show the user posing two questions on the IDP domain (see \cref{subsubsec:model:IDP}).
The initial question in \cref{fig:userioidp}\,\subref{fig:userioidp-q1}
asks for which the situations \textit{typically} result in the NN
outputting a low torque and high state value estimate (Line 1).
In order to produce an answer, Fanoos (Lines 2--3) asks for a preference of initial refinement granularity (
given relative to $S_I$'s side lengths; $\epsilon$ in \cref{eq:stopFunct}),
 and after the user requests $0.125$ (Line 4), lists several potential situations (Lines 5--13).
The user wants more details, and so requests a less abstract description (Line 16); Fanoos now responds with 18 more detailed
situation descriptions (5 listed in
\cref{fig:userioidp}\,\subref{fig:userioidp-a1}, Lines 17--23).
In the second question in \cref{fig:userioidp}\,\subref{fig:userioidp-q2}, the user (Line 25) wants to know the circumstances in which the learned component outputs a high torque while its inputs (e.g., sensors) indicate that the first pole has a low rotational speed; Fanoos finds 32 descriptions (5 listed, Lines 26--34). The user requests a more abstract summary (Line 35), which condenses the explanation down to 3 situations (Lines 36--40).
We see that in both cases---the first request for less abstractness, and the second
for greater---that the explanations adjusted as one would expect, both with respect to the verbosity of the descriptions returned and the verbiage used.

For comments relating to aesthetic extensions to Fanoos's UI, see \cref{subsec:FanoosAndPrettyUI}.

Results generated in February 2020 by code with git-hash \seqsplit{d01c48542f53d1b55376a383a8d50ddd97e59837};
we might include extensions / updates to analysis and UI examples at a later date.

\begin{figure}[t!]
\begin{minipage}{.45\textwidth}
\begin{subfloat}[Initial question response, followed by request for less abstract explanation\label{fig:userioidp-q1}]{
    \centering
    \begin{tikzpicture}
    \node[text width=.95\columnwidth] (question1) {
\begin{lstlisting}[firstline=1,lastline=3]
(Fanoos) when_do_you_usually and(outputtorque_low , statevalueestimate_high )?
Enter a fraction of the universe box length to limit refinement to at the beginning.
 Value must be a positive real number less than or equal to one.
\end{lstlisting}
    };
    \node[below=.5cm of question1,text width=.95\columnwidth] (response1) {
\begin{lstlisting}[firstnumber=5,linerange={5-13,16-17}]
5 of 6 lines to print shown. Press enter to show more. Hit ctrl+C or enter letter q to break. Hit a to list all.
=========
//Description:
(0.45789160, 0.61440409, 'x Near Normal Levels')
(0.31030792, 0.51991449, 'pole2Angle_rateOfChange Near Normal Levels')
(0.12008841, 0.37943400, 'pole1Angle_rateOfChange High')
(0.06128723, 0.22426058, 'pole2Angle Low')
(0.02395519, 0.13633780, 'vx Low')a
(0.01147175, 0.01359231, 'pole1Angle Low')
type letter followed by enter key: b - break and ask a different question,
 l - less abstract , m - more abstract, h - history travel
\end{lstlisting}
    };
    \node[below=.5cm of response1,text width=.95\columnwidth] (end1) {};
    \path (question1.south) -- (question1.south east) coordinate[pos=0.3] (question1-mid-se);
    \path (response1.north) -- (response1.north east) coordinate[pos=0.3] (response1-mid-ne);
    \path (response1.south) -- (response1.south east) coordinate[pos=0.3] (response1-mid-se);
    \path (end1.north) -- (end1.north east) coordinate[pos=0.3] (end1-mid-ne);
    \draw[->] (question1-mid-se) --node[left]{\scriptsize{User requests box length}} node[right,text width=1cm] { 
\begin{lstlisting}[firstnumber=4,numbers=right,firstline=4,lastline=4]
0.125
\end{lstlisting}
    } (response1-mid-ne);
    \draw[->] (response1-mid-se) --node[left,text width=3.5cm,align=right]{\scriptsize{User requests less abstract,\\[-1ex]continue at (b)}} node[right,text width=1cm] { 
\begin{lstlisting}[firstnumber=16,numbers=right,numbersep=2pt,
        firstline=18,
        lastline=18]
l
\end{lstlisting}
    } (end1-mid-ne);
    \end{tikzpicture}
}
\end{subfloat}
\end{minipage}
\qquad
\begin{minipage}{.45\textwidth}
\begin{subfloat}[Less abstract explanation, user satisfied, continues with different question\label{fig:userioidp-a1}]{
    \centering
    \begin{tikzpicture}
    \node[text width=.95\columnwidth] (answer2) {
\begin{lstlisting}[firstnumber=17,linerange={19-25}]
5 of 18 lines to print shown. Press enter to//[...]
=========
(0.16153820, 0.31093854, 'And(endOfPole2_x Near Normal Levels, pole1Angle Low, pole1Angle_rateOfChange High, pole2Angle Near Normal Levels, pole2Angle_rateOfChange High, x High)')
(0.14268581, 0.18653883, 'And(endOfPole2_x Near Normal Levels, pole1Angle Low, pole1Angle_rateOfChange High, pole2Angle Near Normal Levels, pole2Angle_rateOfChange Near Normal Levels, x High)')
(0.11771033, 0.12043966, 'And(pole1Angle Near Normal Levels, pole1Angle_rateOfChange Near Normal Levels, pole2Angle High, pole2Angle_rateOfChange Low, vx Low)')
(0.06948142, 0.07269412, 'And(pole1Angle High, pole1Angle_rateOfChange Near Normal Levels, pole2Angle_rateOfChange High, vx Low, x Near Normal Levels)')
(0.04513659, 0.06282974, 'And(endOfPole2_x Near Normal Levels, pole1Angle Low, pole1Angle_rateOfChange High, pole2Angle High, pole2Angle_rateOfChange Near Normal Levels, x High)')q
\end{lstlisting}
    };
    \node[below=.5cm of answer2,text width=.95\columnwidth] (question2) {};
    \path (answer2.south) -- (answer2.south east) coordinate[pos=0.3] (answer2-mid-se);
    \path (question2.north) -- (question2.north east) coordinate[pos=0.3] (question2-mid-ne);
    \draw[->] (answer2-mid-se) --node[left] {\scriptsize{User break, continue at (c)}} node[right,text width=1cm] { 
\begin{lstlisting}[firstnumber=24,
        numbers=right,numbersep=2pt,aboveskip=0pt,belowskip=0pt,
        firstline=28,
        lastline=28]
b
\end{lstlisting}
    }  (question2-mid-ne);
    \end{tikzpicture}
}
\end{subfloat}
\end{minipage}
\\
\begin{subfloat}[Next question, initial response, and user request to make more abstract\label{fig:userioidp-q2}]{
    \centering
    \begin{tikzpicture}
    \node[text width=.95\columnwidth] (question3) {
\begin{lstlisting}[label={lst:idpquestion},firstnumber=25,linerange={32-32}]
(Fanoos) what_are_the_circumstances_in_which and( pole1angle_rateofchange_low__magnitude , outputtorque_high__magnitude )?
\end{lstlisting}
    };
    \node[below=.3cm of question3,text width=.95\columnwidth] (answer3a) {
\begin{lstlisting}[firstnumber=26,linerange={33-41}]
5 of 32 lines to print shown. Press enter to//[...]
=========
(0.12099418, 0.18835537, 'pole2angle_rateofchange_high__magnitude')
(0.10147897, 0.17831770, 'And(pole1angle_on_the_left, pole2angle_on_the_left, pole2angle_rateofchange_low__magnitude)')
(0.09885232, 0.16335186, 'And(pole1angle_on_the_left, pole2angle_on_the_left, pole2angle_turning_counterclockwise)')
(0.07900125, 0.14467123, 'And(pole1angle_on_the_right, pole2angle_on_the_right, pole2angle_turning_clockwise)')
(0.06693577, 0.12822191, 'And(pole1angle_down, pole2angle_to_right, statevalueestimate_very_low)')q
type letter followed by enter key: b - break and ask a different question,
 l - less abstract , m - more abstract, h - history travel
\end{lstlisting}
    };
    \node[below=.3cm of answer3a,text width=.95\columnwidth] (answer3b) {
\begin{lstlisting}[firstnumber=36,linerange={43-47}]
3 of 3 lines to print shown.
=========
(0.44378316, 0.48588134, 'pole2 not near target position') 
(0.33605014, 0.36551887, 'pole2angle_rateofchange_high__magnitude')
(0.22016670, 0.23739381, 'And(pole2angle_to_right, statevalueestimate_very_low)')
\end{lstlisting}
    };
    \draw[->] (question3) --node[left]{\scriptsize{Fanoos answers}} (answer3a);
    \draw[->] (answer3a) --node[left]{\scriptsize{User requests more abstract}} node[right,text width=1cm] { 
\begin{lstlisting}[firstnumber=35,
        numbers=right,numbersep=2pt,aboveskip=0pt,belowskip=0pt,
            firstline=42,
            lastline=42]
m
\end{lstlisting}
    } (answer3b);
    \end{tikzpicture}
}
\end{subfloat}
\caption{Fanoos user session on the inverted double pendulum example}
\label{fig:userioidp}
\end{figure}

\section{Input-Space Bounding Box Values}
\label{sec:appendix:boundingBoxes}
In this section, we list the input-space bounding boxes 
used in our experiments. We list the values here up to
four significant figures. Listings with further precision 
can be found in the code bases.

\begin{table}[h!]
\caption{Inverted double pendulum input-space boxes}
\label{tab:invertedDoublePendulumBB}
\begin{tabularx}{\columnwidth}{Xrr}
\toprule
Variable name & Lower bound & Upper bound \\ \midrule
$x$ & -1 &  1\\
$vx$ & -0.8 & 0.8 \\
$\textit{pole2\_endpoint}$\parnote{$\textit{pole2\_endpoint}$ is a delta-value with respect to $x$. That is,
   in the observation given to the model to standardize, we add $x$ to the value reported for 
$\textit{pole2\_endpoint}$. This choice is motivated by the fact that the model was trained on 
the $\textit{pole2\_endpoint}$ position being measured in free-space, despite the fact that sensible values for this in an observation are highly dependent on $x$, the horizontal position of the cart's center.} & -0.5 & 0.5\\
$\textit{pole1angle}$ & -0.2 & 0.2\\
$\textit{pole1angle\_rateOfChange}$ & -0.6 & 0.6\\
$\textit{pole2angle}$ & -0.04 & 0.04\\
$\textit{pole2angle\_rateOfChange}$ & -0.7 & 0.7\\\bottomrule
\end{tabularx}
\parnotes
\end{table}

\begin{table}[h!]
\caption{CPU Usage input-space boxes}
\label{tab:CPUUsageBB}
\begin{tabularx}{\columnwidth}{Xrr}
\toprule
Variable name & Lower bound & Upper bound \\ \midrule
$\textit{lread}$ & 0.0 & 0.0369\\
$\textit{scall}$ & 0.0095 & 0.4245\\
$\textit{sread}$ & 0.0028 & 0.0992\\
$\textit{freemem}$ & 0.0061 & 0.6275\\
$\textit{freeswap}$ & 0.4324 & 0.8318\\ \bottomrule
\end{tabularx}
\end{table}

\section{Pseudo-Code for Specific-Selection Subroutine}
\label{sec:appendix:psuedocode}

\setHMarginsBeforePeusdocode
\begin{algorithm}
\caption{Find Most Specific Consistent Predicates. Note here that the notation $\{list\}_{index}$
is sequence-access notation, i.e, $\{< a_i|i \in [n]>\}_{j} = a_j$}
\label{mostSpecificPreds}
\SetKwData{Left}{left}\SetKwData{This}{this}\SetKwData{Up}{up}
\SetKwFunction{Union}{Union}\SetKwFunction{FindCompress}{FindCompress}
\SetKwInOut{Input}{input}\SetKwInOut{Output}{output}

\Input{box to fit, $b$; number of random samples to try, $n$; list of predicates to try, $P$; 
       a list of strictly increase real numbers of length starting with $1.0$, $\ell$}
\Output{A set of indices into $P$ of the most specific predicates, $s$}
\BlankLine
$s \leftarrow \{\} $\;
$dimsCovered \leftarrow \{\}$\;
$bCenter \leftarrow getBoxCenter(b)$\;
$bDim = getDimension(b)$\;
\For{$i\leftarrow 0$ \KwTo $\text{length}(\ell) - 1$}{
    $lowerR \leftarrow \{\ell\}_{i}$\; 
    $upperR \leftarrow \{\ell\}_{(i+1)}$\;
	$innerB \leftarrow ((b - bCenter(b)) \times lowerR)  + bCenter(b)$\;
    $outterB \leftarrow ((b - bCenter(b)) \times upperR)  + bCenter(b)$\;
    $randomSamples \leftarrow \{\}$\;
    \For{$j \leftarrow 1$ \KwTo $n$}{
	 $randomSamples \leftarrow randomSamples \cup \{ getRandVecBetweenBoxes(innerBox, outterBox)\}$\;
    }
    \For{$pIndex \leftarrow 0$ \KwTo $\text{length}(P)$}{
        \If{$pIndex \in s$}{
	    continue\;
	}
	\ElseIf{$\text{freeVars}(\{P\}_{pIndex}) \subseteq dimsCovered$}{
	    continue\;
	}
	\ForEach{$v \in randomSamples$}{
	    \tcc{We evaluate the predicate at index pIndex on v to see if it returns false}
	    \If{$\lnot (\{P\}_{pIndex}).eval(v)$}{
                $s \leftarrow s \cup \{pIndex\}$\;
                $dimsCovered \leftarrow dimsCovered \cup \text{freeVars}(P[pIndex])$\;
		break\;
            }
	}
    }
    \If{$\text{length}(dimsCovered) == bDim$}{
	 return $s$\;
    }
}
return $s$\;
\end{algorithm}
\restoreHMarginsAfterPeusdocode

Pseudo-code for our method of finding the most-specific 
conditions for a box are in Algorithm \ref{mostSpecificPreds}. 
In our code, we used $\ell = c \exp(\alpha \times c)$ where
\begingroup
\setlength{\abovedisplayskip}{0pt}
\setlength{\belowdisplayskip}{0pt}
\setlength{\abovedisplayshortskip}{0pt}
\setlength{\belowdisplayshortskip}{0pt}
\begin{align*}
c = [1.0, 1.01, 1.05, 1.1, 1.2, 1.4, 1.8, 2.6]
\end{align*}
\endgroup
 and $\alpha$ 
is a non-negative real-valued parameter we store and manipulate in the state.
Similarly, $n$ is stored in the states.

\section{Further Technical Descriptions and Details}
\label{sec:appendix:moretechdetails}

\pagestyle{empty} %

\subsection{Neural Networks, \AIDs, and Propagating Boxes Through Networks}
\label{sec:networksAndAIDs}

In this subsection, we discuss how we conduct our \AIDsFull~(\AIDs) analysis as applied to a
FFNN. A crucial 
take-away from this discussion is that this process allows us to produce guarantees
over uncountably infinite many elements  
by considering a
finite collection of abstractions.\\

The abstract domains we consider for our \AIDs~ are boxes (a.k.a, hyper-cubes). 
Boxes facilitate a basic implementation since they are easy to manipulate
and easy to produce for bounding outputs, in contrast to more complex convex polytopes 
that would require less trivial constraint solving to check membership and optimize over. This ease of manipulation comes at the
cost of boxes being ``less precise'' than more sophisticated alternatives that have smaller volume while
containing the same critical elements; larger sets suggest that certain relationships among variables
may be feasible when in reality such combinations of values could not occur.\footnote{
In the case of interval arithmetic, this over-approximation and inclusion of additional elements is often called 
the ''wrapping effect'' \cite{kearfott1996interval}.} 
Our approach is not dependent on this choice ---
it has been adopted solely to facilitate a basic implementation --- and thus engineering improvements are welcome
to choose different \AIDs\ if deemed beneficial to our pipeline (see \cref{subsec:supplementalreachability} for further pointers on engineering improvements).
As a separate note, observe that
the process we will describe is related to local
reachability analysis,  since we compute the image of an input set under a policy $\pi$.\\

The process described here leverages the fact that we are dealing with a pre-trained, fixed-weight
FFNN, that has a typical 
MLP-like structure --- namely, that the network consists of layers of units, each unit 
being comprised of a scalar-valued affine transformation of a subset of the previous layer's activations
that is then passed through a non-decreasing (and typically non-linear) activation function to form the final output.
In the case of recurrent neural nets, or other systems with loops, more sophisticated
mechanisms --- such as reachable-set fixed-point calculations (as discussed in \cite{cousot1977abstract}, for instance) --- would be necessary in general.\\

In order to propagate boxes through the network, we need to examine how to transform a box once it 
passes through a unit in the network. That is, if $u : \mathbb{R}^{I_u} \rightarrow \mathbb{R}^{O_u}$ is a unit of the network with input
dimension $I_u$ and output dimension $O_u$, and
$\times_{i \in [I_u]}[a_i, b_i]$
is the input box,\footnote{Cartesian product of closed, real intervals $\closedint{a_i}{b_i}$, and where $\firstn{n} = \{k \in \mathbb{N} \setminus \{0\} \mid k \leq n\}$.}
we need to calculate $u(\times_{i \in [I_u]}[a_i, b_i])$.\\

For ease of discussion, we suppose that the network in question only has one type of activation function, $\rho$, which is non-decreasing ; our reasoning  trivially extends to encompass networks with multiple such functions. 
For a unit $u$ in the network, let $w \in \mathbb{R}^{I_u}$ be the 
weights of the unit, $\beta \in \mathbb{R}$ be the bias, and $x \in \mathbb{R}^{I_u}$ be the input value. We have that:\\
$$u_{linear}(x) = <w, x> + \beta$$
$$u_{\rho}(x) = \rho(<w, x> + \beta) = \rho(u_{linear}(x))$$
here, $<.,.>$ is the $L^2$ inner product. Notice that since $\rho$ is a non-decreasing function, the
extrema of $u_{\rho}(x)$ and $u_{linear}(x)$ occur at the same arguments.
To find the inputs that produce the extreme values of the activation function over the input space, it thus
suffices to find the values in $\times_{i \in [I_u]}[a_i, b_i]$ that maximize or minimize $<w, x>$. Trivial algebra show that:
$$\underset{x \in \times_{i \in [I_u]}[a_i, b_i]}{argmin}<x,w> = <b_i\mathbbm{1}(\{w\}_i \le 0) + a_i\mathbbm{1}(\{w\}_i > 0) | i \in [I_u]>$$
$$\underset{x \in \times_{i \in [I_u]}[a_i, b_i]}{argmax}<x,w> = <a_i\mathbbm{1}(\{w\}_i \le 0) + b_i\mathbbm{1}(\{w\}_i > 0) | i \in [I_u]>$$
where here, $<z_i|i>$ is sequence construction notation,
$\ithval{\cdot}{i}$ accesses the i-th component of a vector, and $\indfun{\cdot}$ is an indicator function ($\indfun{\top} = 1,~\indfun{\bot} = 0$).
With this, we can compute the images of the input space under the activation functions as follows: for $u \in \{u_{linear}, u_{\rho}\}$,
$$u(\times_{i \in I_u}[a_i, b_i]) = [u(\underset{x \in \times_{i \in [I_u]}[a_i, b_i]}{argmin}<x,w>), u(\underset{x \in \times_{i \in [I_u]}[a_i, b_i]}{argmax}<x,w>)]$$ \\

Having established how a box should be propagated through a unit in the network, 
propagation through the entire network follows immediately. Let $u_{i,j}$ be the $i^{th}$ unit on the $j^{th}$ layer, $M_j$ be the $j^{th}$ layer's size,
$\mathscr{I}_{i,j}$ be the input box to unit $u_{i,j}$, and
$m_{i,j} \subseteq \firstn{M_j}$ s.t. $|m_{i,j}| = I_{u_{i,j}}$: we simply feed the output box from one layer into the 
next similar to the usual feed-forward operation: 
\begin{equation}\label{eq:approxImage} 
u_{i,j+1}(\mathscr{I}_{i,j+1}) = u_{i,j+1}\bigl(\bigtimes\nolimits_{h \in m_{i,j+1}   }u_{h,j}(\mathscr{I}_{h,j})\bigr) \enspace \
\end{equation}
Finally, induction shows that these arguments together establish that this process 
produces a set which contains the image of the network over the box. 
Notice that it is during this recursive process that approximation creeps in; consider, for instance,
the bounding rectangle formed for 
a NN with a 2-d inner-layer whose 
output 
exists on a diagonal line 
whenever the network processes instances from the input space. %

Various extensions to this line of reasoning exist. To close this subsection, we briefly discuss one
example that may appear in a pre-processing or post-processing function. For instance, suppose
we are given a network that is trained to output an action --- but that the action must be properly
post-processed by a function \textbf{properlyTransformAction} that
scales and clips the output prior to finally passing the command to actuators. 
As often happens in practice, suppose that \textbf{properlyTransformAction} 
preserves partial orderings of vector inputs,
so the lexical ordering of the outputs is never the %
reverse of the lexical ordering of the inputs (i.e., if $\forall i \in [dim(\vecArrow{x_{1}})]. \{\vecArrow{x_{1}}\}_i \le \{\vecArrow{x_{2}}\}_i$ then
$\forall i \in [dim(\textbf{properlyTransformAction}(\vecArrow{x_{1}}))]. \{\textbf{properlyTransformAction}(\vecArrow{x_{1}})\}_i \le \{\textbf{properlyTransformAction}(\vecArrow{x_{2}})\}_i$). 
It is easy to see that this gives:
$$\Big( \forall w \in \textbf{properlyTransformAction}(\times_{i \in [I_u]}[a_i, b_i]).$$
$$\big( \forall h \in [dim(w)]. \{\textbf{properlyTransformAction}(<a_i | i \in [I_u]>)\}_{h} \le \{w\}_{h}$$
$$\le
			  \{\textbf{properlyTransformAction}(<b_i | i \in [I_u]>)\}_{h}.~~\big)~~\Big)$$
or, put another way, if 
$$A = \textbf{properlyTransformAction}(<a_i | i \in [I_u]>)$$
and 
$$B = \textbf{properlyTransformAction}(<b_i | i \in [I_u]>)$$,
then we have: 
$$ \textbf{properlyTransformAction}(\times_{i \in [I_u]}[a_i, b_i]) \subseteq \times_{i\in [dim(A)]}[\{A\}_i, \{B\}_i]$$

\subsection{Our  \CEGAR-Inspired Process}
\label{sec:CEGARDescription}

In the model-checking world, \CEGARFull~(\CEGAR)~ \cite{clarke2000counterexample} is a well-regarded technique
for soundly ensuring a system meets desirable properties. 
In short, the approach uses intelligently adapted abstract states uncovered through
guided trial and error to attempt verification or refutation; 
if the desirable property
cannot be proven, the algorithm iteratively refines the abstraction based on where the property is in 
doubt, stopping when the property is either provable or has been disproven by a discovered counterexample.
When applied to certain families of discrete programs, results returned by \CEGAR~ are both sound 
and complete --- this, however, comes at the cost of there not generally being a termination 
guarantee for \CEGAR~ unless one is willing to allow sufficient approximations. In practice, approximations 
used with \CEGAR~ tend to err on the side of safety--- that if \CEGAR~  indicates a property
holds, then it is true, but the converse might not hold. This flexibility has allowed for extensions
of the technique to many domains, including in hybrid system analysis \cite{clarke2003verification},
where the state space is necessarily uncountably infinite and system dynamics do not typically have 
exact numerical representations.

We now overview the \CEGAR-like technique we implemented, using the abstraction domain
described in \cref{sec:networksAndAIDs} as a base. As before, we let $\pi$ be a learned system
$\pi : \mathscr{I} \rightarrow \mathbb{R}^{\mathscr{O}_{\pi}}$, where 
$\mathscr{I} \subseteq \mathbb{R}^{I_{\pi}}$ is the  box $\times_{i \in [I_{\pi}]}[a_{\pi, i}, b_{\pi, i}]$
specifying the input space.\footnote{Or a superset of the input space .}
Let $\phi : \mathbb{R}^{I_{\pi}} \times  \mathbb{R}^{\mathscr{O}_{\pi}} \rightarrow \{\top, \bot\}$
be a formula which we would like to characterize $\pi$'s conformance to over $\mathscr{I}$ 
(i.e., find $\{(w, y) \in \mathscr{I} \times \mathbb{R}^{\mathscr{O}_{\pi}} | \phi(w, y) \land (y = \pi(w))\}$  ).
Notice that $\phi$ need-not use all of its arguments --- so, for instance, the value of $\phi$ might only vary with
changes to input-space variables, thus specifying conditions over the input space but none over the output space.
Since \CEGAR~ is not generally guaranteed to terminate (and certainly would not in our \CEGAR-like implementation),
we introduce a function ${\bf STOP}: \mathbb{R}^{\mathscr{I}_{\pi}} \rightarrow \{\top, \bot\}$
which will be used to prevent unbounded depth exploration when determining whether counterexamples are
spurious or not. The outline of the basic algorithm is shown in algorithm \ref{algo:cegarLikeAnalysis}.
As we will discuss in a moment, some steps that typical \CEGAR~ might not be present, due to our slightly
different aims.

The algorithm begins by forming an initial abstraction of the input space.
In our implementation, the initial abstraction does not leverage any expert impressions
as to what starting sets would be informative for the circumstances; instead, we
opted for the simple, broadly-applicable
strategy of forming  high-dimensional
``quadrants'': $2^{I_{\pi}}$ hyper-cubes formed by bisecting the input space along each of its axes;
we could have just as easily used the universal bounding box undivided as the starting box.
The algorithm takes an input-abstraction, $w$, that has yet to be tried and generates a
abstract state, $\text{approxOutputB}$, that contains $\pi(w)$.\footnote{Notice
here that $w$ and $\pi(w)$ are both sets.} If no member of $\text{w}\times\text{approxOutputB}$ is of interest (i.e.,
meets the condition specified by $\phi$), the algorithm returns the empty set. On the other hand, if 
$\text{w}\times\text{approxOutputB}$
has the potential to contain elements of interest then the algorithm continues, attempting to find the
smallest allowed abstract states that potentially include interesting elements. In general, further
examination is performed by refining the input abstraction, then recursing on the refinements; for efficiency,
we also check whether the entire abstract state satisfies $\phi$, in which we are free to partition it into smaller
abstractions without further checks.

While much of our process is in line with a canonical implementation of \CEGAR~, 
there are 
aspects which we have modified or did not need to implement for our purposes.
For example, in a canonical implementation of \CEGAR, whenever an \AID~is found that potentially
violates the
verification condition,
concrete states within the offending \AID~are then examined
in order to determine if the violation is spurious (that is, a result of wrapping/approximation
effects from working in the abstract space, as oppossed to behaviour exhibited by the concrete system).
Often, if a concrete counterexample is found, the analysis simply stops, but if it is spurious, 
the \AID~ is refined. This arrangement does not necessarly fit our use case, however.
While our process may be analogous or mappable to \CEGAR~ and its standard extensions,\footnote{
For example, sampling-based feasibility checks we perform prior to calling a sat-checker on 
$\text{verdict}_1$ and $\text{verdict}_2$ in algorithm \ref{algo:cegarLikeAnalysis} may be comparable to the spuriousness check.
} we will
refer to our exact approach as a \CEGAR-like process from here-on in order to avoid confusion 
over details or suggest a commitment to stringent canon adherence (e.g., using \cite{clarke2000counterexample} verbatim).

Currently, our method of refinement is to split the box in question along 
the longest ``scaled'' axis. Rigorously, given a box to refine, $\times_{i \in [I_{\pi}]}[a'_{i}, b'_{i}]$, we form
$k$-many new boxes as follows:
$$h = \underset{i \in [I_{\pi}]}{argmax}\frac{b'_{i} - a'_{i}}{b_{\pi, i} - a_{\pi, i}}$$
$$C_k = \frac{b'_{h} - a'_{h}}{k}$$
$${\bf refine}_k(\times_{i \in [I_{\pi}]}[a'_{i}, b'_{i}]) = 
    \cup_{j=0}^{k-1}\Big\{\times_{i \in [I_{\pi}]}[~~a'_{i} +  \mathbbm{1}(i = h)jC_k,~~ 
                                                   b'_{i} +  \mathbbm{1}(i = h)(j +1 -k) C_k~~]\Big\}$$
In our current implementation, we select $k$ at random each time we refine, selecting $k=2$ with probability $0.8$ and 
$k=3$ with probability $0.2$; this was motivated out of an attempt to 
balance between the desires for faster 
\CEGAR-like analysis, further exploration of diverse abstract states, and
keeping the boxes used of sensible size.
The use of $b_{\pi, i} - a_{\pi, i}$ in the denominator for $h$ 
is an attempt to control for differences in scaling and meaning among the variables 
comprising the input space. For instance, 20 millimeters is not commiserate with 20 radians, and our sensitivity to 3 centimeters 
of difference may be different given a quality that is typically on par of kilometers versus one never exceeding a decimeter.

Our refinement strategy allows for efficient caching and reloading of refinement
results by storing the refinement paths, as oppossed to encoding entire boxes. 
Parameters in the state determine if cached results are reused; reuse improves efficiency and 
may help reduce uncalled for volitility in descriptions reported to users, while regenerating results may
produce different \AIDs\ which could lead to a better outcome.

Our analysis used the following {\bf STOP} function, motivated for similar reasons as {\bf refine} from \cref{sec:networksAndAIDs}:
\begin{equation}
\text{{\bf STOP}}(\times_{i \in [k]}[a'_{i}, b'_{i}]) = (~~(\underset{i \in [I_{\pi}]}{max}\frac{b'_{i} - a'_{i}}{b_{\pi, i} - a_{\pi, i}}) \le \epsilon~~) 
\label{eq:stopFunct}
\end{equation}
Here, $\epsilon$ is the refinement parameter initially specified by the user, but which is then automatically
adjusted by operators as the user interactions proceed.

Similar to the choice of \AID, our approach is amenable to more sophisticated refinement and stopping strategies than presented here.

The pseudocode in algorithm \ref{algo:cegarLikeAnalysis} shows the process for the formally sound
questions types. For the probabilistic questions types (i.e., those denoted with ''...usually...''),
$\text{verdict}_1$ is determined by repeated random sampling, and $\text{verdict}_2$ is fixed as $\bot$.
In our implementation, feasibility checks are done prior to calling the SAT solver when handling
a formally sound question-type; we spare  
a through description of efficiency-related modifications for the
sake of presentation clarity.

\setHMarginsBeforePeusdocode
\begin{algorithm}
\caption{Pseudocode for our \CEGAR-like abstract state refinement. Here, inputB is a \AID~ element over the input space (i.e.,
	$\text{inputB} \subseteq \mathscr{I}$). }
\label{algo:cegarLikeAnalysis}

    \SetKwProg{Fn}{Function}{:}{}
    \SetKwFunction{CEGARLikeAnalysis}{CEGARLikeAnalysis}

    \Fn{\CEGARLikeAnalysis{$\text{inputB}$ , {\bf STOP}, $\phi$, $\pi$}}{

        $\text{approxOutputB} \leftarrow \text{approxImage}_{\pi}$(inputB);\tcp{\AIDs-based image approx.; see \cref{eq:approxImage}}
        $\text{verdict}_1 \leftarrow \text{satSolverCheck}(\forall x \in inputB \times \text{approxOutputB}. \lnot \phi(x).)$\;
        \If{$\text{verdict}_1$}{
            \Return $\{\}$ \;
        }
        \If{$\text{\bf STOP}(\text{inputB})$}{
	    \Return $\{\text{inputB}\}$\;
	}
        $\text{verdict}_2 \leftarrow \text{satSolverCheck}(\forall x \in inputB \times \text{approxOutputB}. \phi(x).)$ \;
        \If{$\text{verdict}_2$}{
            $\text{boxesToRefine} \leftarrow \{ \text{inputB} \}$\;
	    $\text{boxesToReturn} \leftarrow \{\}$\;
	    \While{$|\text{boxesToRefine}| > 0$}{
                $\text{thisB} \leftarrow \text{boxesToRefine}.pop()$\;
		\If{$\text{\bf STOP}(\text{thisB})$}{
		    $\text{boxesToReturn} \leftarrow \text{boxesToReturn} \cup \{\text{thisB}\}$\;
		}
		\Else{
                    $\text{boxesToRefine} \leftarrow \text{boxesToRefine} \cup {\bf refine}(\text{thisB})$\;
		}
            }
	    \Return $\text{boxesToReturn}$\;
        }
        $\text{refinedBoxes} \leftarrow \text{\bf refine}(\text{inputB})$\;
	\Return $\cup_{b \in \text{refinedBoxes}}$\CEGARLikeAnalysis{b , {\bf STOP}, $\phi$, $\pi$}\;
    }
\end{algorithm}
\restoreHMarginsAfterPeusdocode

\subsection{Further Details on Automatic Predicate Filter Operator Currently Implemented}
\label{subsec:automatedPredicateSelectionDetails}
\vspace{-10pt}
In this subsection, we provide a slightly more thorough description of the optional operator
implemented in Fanoos which, when utilized, 
automatically determines a predicate to remove from consideration while forming a new description.
For ease of statement, we will refer to the operator in question as APS (for ``automated predicate selector'') in this 
subsection.

Let $S_T$ be the state
whose description, $D_T$, the user currently wants altered.  Let $Q(s)$ be the 
specific question instance\footnote{Here, if the same question is asked later, it is considered
a different instance.} 
for which, in the process of producing replies, a state $s$
was generated.
To determine which named predicate occurring in $D_T$ to remove, 
APS examines records of previous interactions to select a candidate that best 
balances exploration with exploitation.\footnote{
Exploration: trying the available options often enough to be informed of each potential outcome;
Exploitation: choosing the option that will most likely result in the outcome the user requested --- changing the abstraction
level in the desired direction.}
Given a state that occurred in the past, $S_t$,
let:
\begin{itemize}[nosep]
    \item{ $\omega(S_t, p)$ be the number of times a named predicate, $p$, occurs in the description of state $S_t$. This may be greater than one if, for instance, p occurs in multiple conjuncts.}
    \item{$rm(S_t)$ and $rl(S_t)$  be predicates indicating that the user requested the description to become, respectively, more abstract and less abstract (rm: ``request more'') }
    \item{$rb(S_t)$ indicate that the user requested to exit the inner QA-loop (i.e., ``b'' in \cref{lst:lessabstract})
    after seeing $S_t$'s description (rb : ``request break'')}
\end{itemize}
Further, let 
$r_T = rm$ and $r_{T+1} = rl$ if the user requested that $D_T$ (the current description)
 become more abstract,
and $r_T = rl, ~r_{T+1} = rm$ if the user requested lower abstraction.
The predicate which APS removes is determined using the index returned by
\begingroup
\setlength{\abovedisplayskip}{0pt}
\setlength{\belowdisplayskip}{0pt}
\setlength{\abovedisplayshortskip}{0pt}
\setlength{\belowdisplayshortskip}{0pt}
\begin{align*}
\text{UCB}(~\big < |\text{occ}(p)| ~ \big | ~ p~\text{occurs in}~D_T \big >,~\big < |\text{succ}(p)| ~\big | ~ p~\text{occurs in}~D_T \big > ~ )
\end{align*}
where UCB is the deterministic Upper Confidence Bound algorithm \cite{DBLP:conf/focs/Auer00} and 
\begin{align*}
\text{occ}(p) &= \{S_t \in \text{history} | r_{T}(S_t) \land (\omega(S_t, p) > \omega(S_{t+1}, p)) \}\\
\text{succ}(p) &= \{S_t \in \text{occurs}(p) | r_{T+1}(S_{t+1}) \lor rb(S_{t+1}) \}
\end{align*}
where ``$S_t \in \text{history}$'' is a slightly informal reference to accessing $S_t$ from \textit{all} previous interaction records
(i.e., not just replies about $Q(S_t)$ or records from this user session).
$S_{t+1}$ indicates the state that followed $S_t$ \textit{while responding to the same question, $Q(S_t)$} (i.e., it is not simply
any state that comes chronologically after $S_t$ in database records); 
in the cases where $S_{t+1}$ does not exist, we substitute infinity for $\omega(S_{t+1}, p)$,
and false for both $r_{T+1}(S_{t+1})$ and  $rb(S_{t+1})$.
\endgroup

While alternatives to the adopted method could be used --- particularly approaches with greater
stochasticity --- we believe our choice of a UCB algorithm is 
most likely
appropriate at this stage, considering data efficiency and the 
likely nature of the environment.\footnote{For instance, we do not expect an
adversarial environment, nor do we expect --- provided the history of states/replies responding to $Q(S_T)$ --- explicit time dependence.}
Future improvements or novel operators may introduce different or more sophisticated methods for
predicate selection.

\subsection{Pseudocode for the Generation Process after Determining Boxes to Describe}
\label{subsec:generateDescription}

In our pseudocode, $\setOfAllBoxes$ is the set of all boxes that are subsets of $\mathscr{I}$ (see
\cref{sec:CEGARDescription}), and $\partialFunc$ denotes a partial function.

Note that this code assumes pass by copy, not pass by reference.

Further, assume any parameters not passed into the function arguments are 
accessed through a globally accessible state.

Finally, as noted in the main write-up, the boxes being described here may have undergone
some merging after discovery (i.e., post-\CEGAR), but prior to reaching here. Such merging
attempts to increase the abstraction level while retaining some finer-details; it is not
the case in general that increasing $\epsilon$ in the {\bf STOP} function from 
\cref{sec:CEGARDescription} would produce the same results. 

\setHMarginsBeforePeusdocode

\begin{algorithm}
\caption{Pseudocode for the description generation after boxes to described have been determined.}
\label{algo:generateDescription}
\startThisEnvCounter

\SetKwProg{Fn}{Function}{:}{}

\SetKwFunction{generateDescription}{generateDescription}

\tcc{Input: Bs is the collection of boxes found by our \CEGAR-esque process after any
desired post-discovery merging, Cs is the collection of conditions (i.e., named predicates
or conjunctions of them) that may be used to produce descriptions, and produceGreaterAbstraction
is a boolean parameter stored in the state that operators may toggle. nSample is an integer
determining the number of random-sample feasibility checks to do prior to calling a SAT solver,
a procedure only to aid efficiency.}
\tcc{Output: a description in DNF form, weights for each conjunct in the DNF formula representing
unique coverage of each conjunct, weights representing the total coverage of each conjunct (including
possible redundancies), and the list of conditions after any additions formed during the description
generation (i.e.,
new conjunctions or box-range predicates) .}
\Fn{\generateDescription{ Bs , nSample , Cs , produceGreaterAbstraction }}{
    \If{ |Bs| == 0}{
        throw exception(''No Situation Corresponds to the Event User Described Occurring'')\;
    }

   ( $\text{Cs}_2$ , csAndBs , bsAndGoodCs ) $\leftarrow$ getInitialListOfConditionsConsistentWithBoxes(
	Bs, nSample, Cs, produceGreaterAbstraction) \;

    coveringCs $\leftarrow$ getApproximateMultivariateSetCover( bsAndGoodCs )\;

    \tcp{The below line is needed because coveringCs may contain new conjuncts.}
    $\text{Cs}_3 \leftarrow \text{Cs}_2 \cup~ \text{coveringCs}$\;

    \BlankLine
    Make: $\text{csToBs}: \text{Cs}_2 \partialFunc \mathscr{B}$ \\
    s.t. $\forall c \in \text{Cs}_2. \forall b \in \mathscr{B}. (~(b \in csToBs(c)) \iff ( (c, b) \in csAndBs)~)$ \;
    \BlankLine

    $\text{csToBs}_2 \leftarrow$  handleNewConjuctions( $\text{Cs}_3$ , csToBs )\;

    \BlankLine
    Make: $\text{bsToCs}_2 : \mathscr{B} \partialFunc coveringCs$\\
    s.t. $\forall b \in \mathscr{B}. \forall c \in coveringCs. (~( c \in \text{bsToCs}_2(b)) \iff (b \in \text{csToBs}_2(c)~))$\;
    \BlankLine

    \tcc{ Below, a second covering is done in order to account for predicates that cover
        a superset of the boxes of another predicate returned in coveringCs. Notice that this would {\it not} necessarily
        have been handled by the previous call to getApproximateMultivariateSetCover 
        since there we only listed which boxes
        a predicate was {\it among the most specific for}, not the the total set of
        boxes it was consistent with. So, for instance, it is possible that a box was only
        describable by one vague predicate in the results from the first covering - this
        covering would handle the fact that the vague predicate may imply the behaviour of
        many of the other specific predicates. See earlier in the write-up for how we avoid
        this ''washing-out'' finer-grained detail. 
    }
    $\text{coveringCs}_2 \leftarrow \text{getApproximateMultivariateSetCover}(\text{bsToCs}_2)$\;

    $\text{Cs}_4 \leftarrow \text{Cs}_3 \cup  \text{coveringCs}_2$\;

    $\text{csToBs}_3 \leftarrow$  handleNewConjuctions( $\text{Cs}_4 ,~\text{csToBs}$ )\;

    $(\text{csToTV} , \text{csToUV}) \leftarrow$ getVolumesCoveredInformation( Bs, $\text{coveringCs}_2$,
        $\text{csToBs}_3$)\;

    $\text{coveringCs}_3 \leftarrow$ removePredicatesImpliedByOthers(
        $\text{coveringCs}_2$, $\text{csToBs}_3$, $\text{csToUV}$ )\;

    ($\text{coveringCs}_4, ~\text{Cs}_5, ~\text{csToBs}_4$) $\leftarrow$ handleNewInstancesOfBoxRangePred($\text{coveringCs}_3, ~\text{Cs}_4, ~\text{csToBs}_3$)

    $(\text{csToTV}_2 , \text{csToUV}_2) \leftarrow$ getVolumesCoveredInformation(
        Bs, $\text{coveringCs}_3, ~\text{csToBs}_4$  )\;

    \tcc{Note that, when presenting results to users, we primarily use elements from $\text{coveringCs}_4$
        to index into other structures. Thus, it is acceptable if other structures happen to have
        domains that are supersets of $\text{coveringCs}_4$.
        }
    \Return ($\text{coveringCs}_4$,
             $\text{Cs}_5$,
	     $\text{csToTV}_2$ ,
	     $\text{csToUV}_2$) \;
} 
\updateGlobalAlgoLineCount
\end{algorithm}

\begin{algorithm}
\caption{Pseudocode for determining which predicates cover boxes and which are most specific.}
\label{algo:getInitialListOfConditionsConsistentWithBoxes}
\startThisEnvCounter

\SetKwProg{Fn}{Function}{:}{}

\SetKwFunction{getInitialListOfConditionsConsistentWithBoxes}{getInitialListOfConditionsConsistentWithBoxes}

\Fn{\getInitialListOfConditionsConsistentWithBoxes{Bs, nSample, Cs, produceGreaterAbstraction}}{
    $\text{Cs}_2 \leftarrow \text{Cs}$\;
    $\text{csAndBs} \leftarrow \{\}$;\tcp{$\text{csAndBs} \subseteq \text{Cs}_2 \times \text{Bs}$}
    $\text{bsAndGoodCs} \leftarrow \{\}$; \tcp{$\text{bsAndGoodCs} \subseteq \text{B} \times \text{Cs}_2$}
    \For{ b $\in$ Bs}{
        CsForThisB $\leftarrow$ getConsistentConditions(b, nSample, Cs)\;
	\For{c $\in$ CsForThisB}{
            csAndBs $\leftarrow$ csAndBs $\cup$ \{(c, b)\}\;
        }
        mostSpecificCsForThisB $\leftarrow$  getMostSpecificCondition(b, nSample, CsForThisB)\;

	\If{ (mostSpecificCsForThisB == Null) $\lor$ (|CsForThisB| == 0) }{ 
	    \If{ ($\lnot$ produceGreaterAbstraction) $\lor$ (|CsForThisB| == 0) }{
                newBoxP $\leftarrow$ createBoxPredicate(b)\;

		$\text{Cs}_2 \leftarrow \text{Cs}_2 \cup$ \{ newBoxP \}\;

                csAndBs $\leftarrow$ csAndBs $\cup$ \{ (newBoxP, b) \}\;

                bsAndGoodCs $\leftarrow$ bsAndGoodCs $\cup$ \{ (b,newBoxP) \}\;
            }
	    \Else{
                bsAndGoodCs $\leftarrow$ bsAndGoodCs $\cup$ \{ (b,c) | c $\in$ CsForThisB \}\;
            }
	}
	\Else{
            bsAndGoodCs $\leftarrow$ bsAndGoodCs $\cup$ \{(b, c) | c $\in$ mostSpecificCsForThisB \}\;
        }
    }
    \Return ( $\text{Cs}_2$ , csAndBs , bsAndGoodCs )\;
} \updateGlobalAlgoLineCount
\end{algorithm}

\begin{algorithm}
\caption{Helper function for algorithm \ref{algo:getInitialListOfConditionsConsistentWithBoxes}}
\label{algo:getConsistentConditions}
\startThisEnvCounter
\SetKwProg{Fn}{Function}{:}{}
\SetKwFunction{getConsistentConditions}{getConsistentConditions}

\Fn{\getConsistentConditions{b, nSamples, ps}}{

    samples $\leftarrow \{\}$\;
    \For{  $i \leftarrow 1; i <= \text{nSamples}; i \leftarrow i + 1$}{
	samples $\leftarrow$  samples $\cup$ \{randomVectorInBox(b)\}\;
    }
    candidatePs = \{\}\;
    \For{thisP $\in$ ps}{
        success $\leftarrow \top$\;
	\For{thisS $\in$ samples}{
	    \If{ $\lnot$ thisP( thisS) }{
                success $\leftarrow \bot$\;
                break\;
	    }
	}
	\If{success}{
            candidatePs $\leftarrow$ candidatePs $\cup$ \{thisP\}\;
	}
    }
    consistentPs $\leftarrow \{\}$\;
    \For{thisP $\in$ candidatePs}{
        verdict $\leftarrow$  satSolverCheck($\forall v \in b. \text{thisP}(v)$)\;
	\If{verdict}{
            consistentPs $\leftarrow$ consistentPs $\cup$ \{thisP\}\;
	}
    }
    \Return consistentPs\;
}
\updateGlobalAlgoLineCount
\end{algorithm}

\begin{algorithm}
\caption{Pseudocode for the multi-dimensional set covering we conduct.}
\label{algo:getApproximateMultivariateSetCoverPartA}
\startThisEnvCounter

\SetKwProg{Fn}{Function}{:}{}

\SetKwFunction{getApproximateMultivariateSetCover}{getApproximateMultivariateSetCover}
\Fn{\getApproximateMultivariateSetCover{bsAndPs}}{
(bsToPs, psToBs) $\leftarrow$ startFunct(bsAndPs) \;
filteredDomainMaxPsToBs $\leftarrow$ getMaxCover(bsToPs, psToBs)\;
setsOfSetOfPsToConjunct $\leftarrow$ reverseOut(filteredDomainMaxPsToBs)\;
conditionList $\leftarrow$ couplePs( setsOfSetOfPsToConjunct ) \;
\Return conditionList \;
}
\BlankLine
\BlankLine

\SetKwFunction{startFunct}{startFunct}

\Fn{\startFunct{bsAndPs}}{
Make: $~~\text{bsToPs} : \mathscr{B} \partialFunc \mathscr{P}( \text{AllPreds}) $ \\
Initailize such that: $~~\forall b \in \mathscr{B}. ~\text{bsToPs}(b) = \{ p \in \text{AllPreds} | (b, p) \in \text{bsAndPs}\}$\;
\BlankLine
Make: $~~\text{psToBs} : \text{AllPreds} \partialFunc \mathscr{P}(\mathscr{B})$\\
Initailize such that: $~~\forall p \in \text{AllPreds}. ~\text{psToBs}(p) = \{ b \in \mathscr{B} | (b, p) \in \text{bsAndPs}\}$\;
\BlankLine
\Return (bsToPs, psToBs) \;
}
\BlankLine
\BlankLine

\SetKwFunction{getMaxCover}{getMaxCover}
\Fn{\getMaxCover{bsToPs, psToBs}}{
    Make: $~~\text{maxPsToBs} : \text{AllPreds} \partialFunc \mathscr{P}(\mathscr{B})$\;
    Initialized so that: $~~\forall p \in \text{AllPreds}.~ \text{maxPsToBs}(p) = \{\}.$\;
    \BlankLine
    Make: $~~\text{bsToCoveredXs} : \mathscr{B} \partialFunc \mathscr{P}(SetOfVariables)$\;
    Initialized so that: $~~\forall b \in \mathscr{B} .~ \text{bsToCoveredXs(b)} = \{\}$\;
    \BlankLine
    \While{|domain(bsToPs)| > 0}{
        maxP $\leftarrow$ Null; \tcp{empty value for now} 
        bsCoveredByMaxP $\leftarrow \{\}$\;
	\For{thisP $\in$ domain(psToBs)}{
             \If{ $( |\text{psToBs}(\text{thisP})| > |\text{bsCoveredByMaxP}| ) \lor$\\
		$(~( |\text{psToBs}(\text{thisP})| == |\text{bsCoveredByMaxP}| ) \land (x > 0.5 ~\text{where} ~x \sim \text{Uniform}([0,1]) )~)$}{
                maxP $\leftarrow$ thisP\;
                bsCoveredByMaxP $\leftarrow$ psToBs(thisP)\;
             }
	}
        maxbsAndPs(maxP) $\leftarrow$ maxbsAndPs(maxP) $\cup$ \{bsCoveredByMaxP\} \;
        xsCoveredByMaxP $\leftarrow$ freeVars(maxP) \;
	\For{thisB $\in$ bsCoveredByMaxP}{
            bsToCoveredXs(thisB) $\leftarrow$ bsToCoveredXs(thisB) $\cup$ xsCoveredByMaxP \;
            psNowCoveringThisB $\leftarrow$ bsToPs(thisB) \;
	    \For{thisP $\in$ psNowCoveringThisB}{
                \If{ $\lnot$ ( freeVars(thisP) $\subseteq$ bsToCoveredXs(thisB) ) }{
                    continue \;
                }
                psToBs(thisP) $\leftarrow$ psToBs(thisP) $\setminus$ \{thisB\}\;
                bsToPs(thisB) $\leftarrow$ bsToPs(thisB) $\setminus$ \{thisP\}\;
	    }
	    \If{ bsToPs(thisB) == \{\}}{
                bsToPs $\leftarrow bsToPs \restriction ( domain(bsToPs) \setminus \{thisB\} )$
	    }
	}
    }
 
    Make: $\text{filteredDomainMaxPsToBs} :~ \text{AllPreds}_2 \partialFunc \mathscr{P}(\mathscr{B}) , ~~\text{AllPreds}_2 \subseteq \text{AllPreds}$\\
    s.t. $\text{filteredDomainMaxPsToBs} = \text{maxPsToBs} \restriction \{ p \in \text{AllPreds} | \text{maxPsToBs}(p) \not= \{\} \}$\;
    \BlankLine
    \Return filteredDomainMaxPsToBs\;
}
\updateGlobalAlgoLineCount
\end{algorithm}

\begin{algorithm}
\caption{Continuation of algorithm \ref{algo:getApproximateMultivariateSetCoverPartA}, helper functions for the multi-dimensional covering process.}
\label{algo:getApproximateMultivariateSetCoverPartB}
\startThisEnvCounter

\SetKwProg{Fn}{Function}{:}{}

\SetKwFunction{reverseOut}{reverseOut}

\Fn{\reverseOut{fDMaxPsToPs}}{
\tcc{Below, each element is exactly a set of predicates used to cover a member of  B.\\
    SSPC stands for ''set of set of predicates to conjunct''}
	$\text{SSPC} \leftarrow \{ ps \in \mathscr{P}(\text{AllPreds}) |$ \\
	$~~~~\exists b \in \text{B}.~\forall p \in \text{domain}(\text{fDMaxPsToBs}). (b \in \text{fDMaxPsToBs}(p) \iff  (p \in ps))$\}\;
    \Return SSPC\;
}

\SetKwFunction{couplePs}{couplePs}
\BlankLine
\BlankLine

\Fn{\couplePs{SSPC}}{
    resultCs  $\leftarrow$ \{\}\;
    \For{thisSetOfPs $\in$ SSPC}{
	 \If{|thisSetOfPs| > 1}{
	 resultCs $\leftarrow \text{resultCs} \cup \{ \land_{(p \in \text{thisSetOfPs})}~p \}$\;
	 }
	 \Else{
             resultCs $\leftarrow \text{resultCs} \cup \text{thisSetOfPs}$\;
         }
    }
    \Return resultCs ;	
}
\updateGlobalAlgoLineCount
\end{algorithm}

\begin{algorithm}
\caption{Pseudocode to properly adjust book-keeping structures after introduction of conjuncts during the covering process.}
\label{algo:handleNewConjuctions}
\startThisEnvCounter
\SetKwProg{Fn}{Function}{:}{}
\tcp{Reminder: this pseudocode does not mutate the caller's copy of arguments}
\SetKwFunction{handleNewConjuctions}{handleNewConjuctions}

\Fn{\handleNewConjuctions{Cs , csToBs}}{
    \For{thisC $\in$ Cs}{
        \If{thisC $\in$ domain(csToBs)}{
	    continue \;
	}
	\tcc{Below, isAConjunct: returns $\top$ IFF thisC is of form $\land_{p \in \text{ps}}p$ for a set of predicates (atomic statements) ps}
        \If{isAConjunct(thisC)}{
           bsCoveredByThisC $\leftarrow$ Null; \tcp{empty value to start}
           \tcp{Below iterate of thisP where $\text{thisC} = \land_{\text{thisP} \in \text{ps}}\text{thisP}$}
           \For{ thisP $\in$ getAtoms(thisC)}{
               bsCoveredByThisP $\leftarrow$ csToBs(thisP)\; 
   	       \If{bsCoveredByThisC == Null}{
	           bsCoveredByThisC $\leftarrow$ bsCoveredByThisP\;	   
               }
	       \Else{
	           bsCoveredByThisC $\leftarrow$ bsCoveredByThisC $\cap$ bsCoveredByThisP\;
	       }
               domain(csToBs) $\leftarrow$ domain(csToBs) $\cup$ \{thisC\}\;
               csToBs(thisC) $\leftarrow$ bsCoveredByThisC\;
	    }
        }
    }
    \Return csToBs \;
} \updateGlobalAlgoLineCount
\end{algorithm}

\begin{algorithm}
\caption{Pseudocode for determining volume information.}
\label{algo:getVolumesCoveredInformation}
\startThisEnvCounter
\SetKwProg{Fn}{Function}{:}{}

\SetKwFunction{getVolumesCoveredInformation}{getVolumesCoveredInformation}

\tcc{Below, ''Conditions'' (typically denoted with a ''c'' in naming) include atomic predicate(
    i.e., named predicates and box-range predicates) and conjunctions of atomic predicate.}
\Fn{\getVolumesCoveredInformation{B, Cs, csToBs}}{

     Make: $\text{csToBs} : \text{Cs} \partialFunc \mathscr{P}(\mathscr{B})$\\
     s.t. $\forall c \in \cup_{b \in \text{domain}(\text{bsToCs})}\text{bsToCs}(b).~ \text{csToBs}(c) = \{b \in \mathscr{B} |c \in  \text{bsToCs}(b) \}$\;
     \BlankLine

     \tcp{For total volume}
     Make: $~~~\text{csToTV} : \text{Cs} \partialFunc \{r \in \mathbb{R} | r >= 0\}$\\
     Initialized so that $\forall c \in \text{Cs}.~\text{cvToTV}(c) = 0.0$\;
     \BlankLine

     \tcc{ Below, UV stands for ''unique volume'',
           the total volume a condition is responsible for uniquely covering. 
           Depending on one's definition, it may be considered an approximation ---
           though sound lower bound --- for the ''unique'' volume, since multiple
           predicates might cover the same box, but different axes (i.e., variables),
           a sort of ''uniqueness'' not accounted for by csToUV . Naturally, the 
           alternate notion could be implemented using trivial, additional bookkeeping;
           this alternative, however, would likely be redundant with csToTV in 
           most cases.}
     Make: $~~~\text{csToUV} : \text{Cs} \partialFunc \{r \in \mathbb{R} | r >= 0\}$\\
     Initialized so that $\forall c \in \text{Cs}.~\text{cvToUV}(c) = 0.0$\;
     \BlankLine 

     TV = 0.0 ; \tcp{total volume of all the boxes}
     \For{thisB $\in$ B}{
        v $\leftarrow$ computeVolume(thisB)\;
        TV $\leftarrow$ TV + v\;
        csCoveringThisB $\leftarrow$ bsToCs(thisB)\;
	\For{$\text{thisC} \in \text{csCoveringThisB}$}{
            csToTV(thisC) $\leftarrow$ csToTV(thisC) + v\;
	    \If{ |csCoveringThisB| == 1}{ 
                csToUV(thisC) $\leftarrow$ csToUV(thisC) + v\;
            }
        }
     }

    \If{TV == 0.0}{
        TV = 1.0\;
    }
    \BlankLine
    Make: $~~~\text{csToTV}_2 : \text{Cs} \partialFunc \{r \in \mathbb{R} | r >= 0\}$\\
    s.t. $\forall c \in \text{domain}(\text{csToTV}).~\text{csToTV}_2(c) = \frac{\text{csToTV}(c)}{\text{TV}}$\;
    \BlankLine

    Make: $~~~ \text{csToUV}_2 : \text{Cs} \partialFunc \{r \in \mathbb{R} | r >= 0\}$\\
    s.t. $\forall c \in \text{domain}(\text{csToUV}).~\text{csToUV}_2(c) = \frac{\text{csToUV}(c)}{\text{TV}}$\; 
    \BlankLine

    \Return ($\text{csToTV}_2$ , $\text{csToUV}_2$)\;
} \updateGlobalAlgoLineCount
\end{algorithm}

\begin{algorithm}
\caption{Pseudocode to remove description elements that are implied by other parts of the description. Note
that a particular predicate may cover a box that no other {\it individual} predicate covers - but the appropriate
disjunction of other predicates might; this is exactly what we check for here. Some code for efficiency evoked
prior to this stage - such as 
sound feasibility checks and faster-but-incomplete redundancy-removal functions (incomplete in the sense that not all
 redundancy can be addressed by them) -
are not shown for clarity.}
\label{algo:removePredicatesImpliedByOthers}
\startThisEnvCounter

\SetKwFunction{removePredicatesImpliedByOthers}{removePredicatesImpliedByOthers}
\SetKwFunction{disjunctCoversBox}{disjunctCoversBox}

\SetKwProg{Fn}{Function}{:}{}

\Fn{\removePredicatesImpliedByOthers{Cs, csTobs, csToUniqVols}}{
    
    Bs $\leftarrow \cup_{\text{c} \in \text{Cs}}$ csTobs(c)\;

    boxCs $\leftarrow \{ \text{c} \in \text{Cs} | \text{c is a box-range predicate}\}$\;
    conjunctionCs $\leftarrow \{\text{c} \in \text{Cs} | \text{c is a conjunction of atomic predicates}\}$\;
    othersCs $\leftarrow \text{Cs} \setminus (\text{boxCs} \cup \text{conjunctionCs} )$\;

    orderToConsider $\leftarrow <>$;\tcp{Note: <> is a sequence}
    \tcc{Below: iterate from left-most (0-index) to right-most (max index), in order}
    \For{s $\in$ <boxCs, conjunctionCs, othersCs>}{
        \tcp{Below: cons(<A, B>, C) evaluates to <A, B, C>, i.e., append to back}
        orderToConsider $\leftarrow$ cons( orderToConsider, 
            formSortedList( listToSort $\leftarrow$ s, sortKey $\leftarrow$ csToUniqVols, 
                order $\leftarrow$ 'ascending') )\;
    }

    $\text{Cs}_2 \leftarrow$ Cs\;
    bsToAlwaysCheck $\leftarrow \{\}$\;
    \tcp{Below: as before, iterate from 0-index to max-index}
    \For{thisList $\in$ orderToConsider}{
        \For{thisC $\in$ thisList}{
            restOfCs $\leftarrow \text{Cs}_2 \setminus \{\text{thisC}\}$\;
            bsToCheckHere $\leftarrow \text{bsToAlwaysCheck} \cup \text{csTobs}(\text{thisC})$\;
            removeThisC $\leftarrow \top$ \;
            \For{thisB $\in$ bsToCheckHere}{
                \If{ $\lnot$ disjunctCoversBox(thisB, restOfCs) }{
                   removeThisC $\leftarrow \bot$ \;
                   break \;
                }
            }
            \If{removeThisC}{
                $\text{Cs}_2 \leftarrow$ restOfCs \;
                bsToAlwaysCheck $\leftarrow$ bsToCheckHere \;
            }
        }
    }

    \Return $\text{Cs}_2$ \;
}
\updateGlobalAlgoLineCount
\end{algorithm}

\begin{algorithm}
\caption{Pseudocode for helper-function to algorithm \ref{algo:removePredicatesImpliedByOthers}}
\label{algo:disjunctCoversBox}
\startThisEnvCounter

\SetKwFunction{disjunctCoversBox}{disjunctCoversBox}

\SetKwProg{Fn}{Function}{:}{}

\Fn{\disjunctCoversBox{b, Cs}}{

    \tcc{first, for efficiency, some random sampling prior to a formal check }
    nSamples $\leftarrow$ configurationFile.nSamplesPriorToFormalCheck \;
    \For{ $i \leftarrow 1; i <= \text{nSamples}; i \leftarrow i +1$}{
        v $\leftarrow$ getRandomVectorInBox(b) \;
        noCHolds $\leftarrow \top$ \; 
        \For{thisC $\in$ Cs}{
            \If{thisC(v)}{
                noCHolds $\leftarrow \bot$\;
                break\;
            }
        }
        \If{noCHolds}{
            \Return $\bot$
        }
    }

    \tcp{ formal check with SAT solver}
    $\text{formulaToCheck} \leftarrow (\forall v \in \text{b}. ~\lor_{\text{thisC} \in \text{Cs}} ~\text{thisC}(v))$\;
    \Return satSolverCheck(formulaToCheck)\;
}

\updateGlobalAlgoLineCount
\end{algorithm}

\begin{algorithm}
\caption{Pseudocode for handling boxes that fail to be described by conditions (named-predicates and conjunctions of
named-predicates) earlier in the process. In particular, we attempt to merge any boxes remaining at this phase prior to
displaying to the user.}
\label{algo:handleNewInstancesOfBoxRangePred}
\startThisEnvCounter
\SetKwProg{Fn}{Function}{:}{}

\SetKwFunction{handleNewInstancesOfBoxRangePred}{handleNewInstancesOfBoxRangePred}

\Fn{\handleNewInstancesOfBoxRangePred{coveringCs, Cs, dictMappingConditionToBoxesItIsConsistentWith}}{

    listOfCandidateBoxes $\leftarrow \{ \text{thisC} \in \text{coveringCs} | \text{thisC is a box-range predicate}\}$\;
    $\text{Cs}_2 \leftarrow  \text{Cs} \setminus \text{listOfCandidateBoxes}$ \;
    $\text{coveringCs}_2 \leftarrow  \text{coveringCs} \setminus \text{listOfCandidateBoxes}$ \;
    newMergedBoxes $\leftarrow$ mergeBoxes( $\cup_{\text{thisC} \in \text{listOfCandidateBoxes}}\{\text{thisC.box}\}~$, state.parametersForBoxMerging)\;

    $\text{newConditionsToAddIn} \leftarrow \cup_{\text{thisBox} \in \text{newMergedBoxes}} \{ \text{createBoxPredicate}( \text{thisBox}) \}$\;
    \For{thisC $\in$ newConditionsToAddIn}{
        $\text{Cs}_2 \leftarrow \text{Cs}_2 \cup \{\text{thisC}\}$\;
	$\text{coveringCs}_2 \leftarrow  \text{coveringCs}_2 \cup \{ \text{thisC} \}$\;
        \tcc{Below is a bit of a sub-preferred and slow way to do things - 
             ideally we would keep track of this information when we did the original
             merges - but as-written, it is reasonable way to write it to get across the idea.
             Further, there would ideally not be too many boxes here, as hopefully most
	     boxes would be addressed by named-predicates. 
	     Either way, the ideal way would be to have
             mergeBoxes returned the merged boxes and a set of tuples
             of form (original box, box in the new merger) to avoid the 
             repeated work of the below line.}
        $\text{dictMappingConditionToBoxesItIsConsistentWith}(\text{thisC}) \leftarrow$
          $ \{ b \in \text{listOfCandidateBoxes} | \text{firstBoxContainsSecondBox}(\text{thisC.box}, b) \}$ \;
     }

    \Return ($\text{coveringCs}_2$, $\text{Cs}_2$, dictMappingConditionToBoxesItIsConsistentWith)\;
} \updateGlobalAlgoLineCount
\end{algorithm}

\restoreHMarginsAfterPeusdocode

\newpage

\section{Extended Comment on the Difficulties of Pretheoretical Notions of Abstractness}
\label{sec:appendix:pretheoreticalNotionsOfAbstractness}
As commented on in \cref{subsec:methodologyCapturingConceptOfAbstractness}, criteria to judge
degree-of-abstractness in the lay sense are often difficult to capture, if even existent generally.
For instance, one may ponder which is more abstract:
a chair or the collection of atoms that constitute it? Does it matter if we consider only  
one atom in the chair as opposed to all the atoms? It is debatable which, if either, should 
be considered more abstract than the other in any absolute sense. The history of science suggests 
one ordering, while a casual, crude consideration of Platonism crossed with parts-of-whole 
relations suggests another way. We spare further consideration of this complexity to stave-off 
sidetracking too far in this work. At the risk of possibly invoking a platitude and sounding as though   
pragmatic answers are the ultimate answer,\footnote{A position we are not explicitly committing
to or against at this stage.} we have already highlighted in \cref{subsec:methodologyCapturingConceptOfAbstractness}
works that do point to some clear
schemas and demonstrate functioning utilizations of {\it implementations of the concept} in 
computer science work. Regrettably, we will leave the discussion here as to: (1) whether a single
notion of abstractness is applicable across all contexts, desires, collections of objects, and aspects of human life, 
(2) whether the chair-versus-atoms example
may be inappropriate due to (a) mixing types or (b) context / subject matter irrelevant for the
situations we consider in this work.

\section{Errata}  
\label{sec:appendix:errata}

\subsection{Errata for Prior ArXiv Releases of this Paper}
\label{subsec:appendix:errata:arxiv}

\subsubsection{April $\text{13}^{\text{th}}$, 2021:}

We have updated the description of the APS operator, now described in
\cref{subsec:automatedPredicateSelectionDetails}.
While not quite incorrect, we recognize that the 
previous
account --- 
which may have been excessively concise
--- could
unintentionally be 
misleading, and did not do sufficient justice overviewing the factors considered.
Realizing the potential for misunderstanding, and believing that some readers
may be interested in this operator in its own right, we have opted to provide a 
more thorough description.

Previously on ArXiv,
we referred to APS
as using a ``roughly na\"{i}ve-Bayes model'' to make its determinations.
This was intended to reflect, for instance, the fact that it does not 
consider the joint occurrences of
predicates or substantial structure of the context.
Further, the signal APS leverages is essentially a simple
success ratio (i.e., $\frac{succ(p)}{occ(p)}$ ) coupled with 
confidence bounds based on the number of trials; the success ratio is analogous to the
simple class probability estimators found in  
discrete na\"{i}ve-Bayes classifiers.
This all said, however, we have come to recognize that this description fails to do justice
in detailing the implementation,
and particularly misses the explicit consideration for balancing exploitation with exploration.
To be clear, the operator's implementation has always used the UCB algorithm,  as
seen in the code present on GitHub.

\subsubsection{March $\text{19}^{\text{th}}$, 2021:}
\label[subsubsection]{subsubsec:appendix:errata:arxiv:d19M3y2021}
As can be seen by comparing\footnote{We suggest the use of the Unix data comparison utility
 named ''diff'' for this purpose.} the ArXiv write-ups from before and after this date, 
a few noteworthy changes have occurred. Some changes incorporate descriptions that were made in 
\cite{IJCAIXAIFanoos2020} (for example, \cref{sec:sup:fanoosextensions}). Other addenda 
describe extensions to Fanoos that were
made after the initial {\it submission}\footnote{Submission, so certainly prior to acceptance
or publication at that venue.} of this work to \cite{IJCAIXAIFanoos2020}. The chief example of this is the 
operator for automatic predicate filtering which we described at a high-level, the code for which will be publicly
available with the rest of Fanoos.
Most additional, substantial extensions to Fanoos we intend to describe in a future paper, as opposed to making further
addenda to this write-up.

Outside the description of novel
extensions, however, we did correct some aspects of the core algorithm's {\it description}.
The chief correction has been in the discussion of the covering process: in this current
write-up, we verbosely describe the two covering 
processes we undertake. In prior 
writings --- due to the pressures of space, time, and having developed the algorithm core
long before --- we attempted to provide a slimmer, mathematically equivalent
description (e.g., the description in \cite{IJCAIXAIFanoos2020}); this description, upon further review, was
not equivalent to what we intended. Regrettably, owing to other constraints, we returned to correct
this erroneous account only now.\footnote{The first submission of this write-up to ArXiv was similar to the content submitted to
\cite{IJCAIXAIFanoos2020}.} We stress that the core algorithm has not been altered since the time of 
\cite{IJCAIXAIFanoos2020}, or even the abstract-submission to \cite{cmuSymposiumOnAISGFanoos2020} which preceded it;
changes to the core description are simply the result of us being more precise and finally taking proper advantage
of the space freedoms which ArXiv provides. We must also note that the box-merging prior-to and after the covering
generation has always been a part of our algorithm ; we spared some mentioning of this detail in prior writings due to space 
limitations and reviewers' ability to access our code.
While perhaps under-reported in writing, 
we referenced box-merging in various materials found at \cref{sec:appendix:furtherMaterials} and their respective
venues. Though not quite
satisfying, in all circumstances, full information was in principle available to reviewers via the code we provided.
Again, we emphasize: no changes in this write-up are a reflection of alterations to the algorithm's core processes.

\subsubsection{\dateUpdatePostVMCAITwentyTwentyTwo : \\} %
\label[subsubsection]{subsubsec:appendix:errata:arxiv:d9M2y2022}

\textit{Regarding an Error in a Previous Citation Used: } In a earlier version of the
write-up provided on ArXiv, a book review of \cite{laird1986universal} was accidentally
cited instead of the book itself. On a sidenote: thanks goes to Jack Mostow, \orcidID{0000-0003-0121-3334} %
who gave away a nice copy of the book in roughly 2016/2017 on the condition that it
go to a good home; this book has been much enjoyed, along other AI-related material from a similar era provided by Dr. Mostow. We express our thanks.\\

\textit{Regarding Correction of Error in \cref{tab:questionDescriptions}:} Unfortunately,
the "restrictions" listed for "what do you do when" questions and "when do you" questions were 
backwards up until the VMCAI 2022 publication (\cite{DBLP:conf/vmcai/BayaniM22}) and this update on ArXiv.
In particular, this error effects
prior versions of this write-up on ArXiv as well as the publication at \cite{IJCAIXAIFanoos2020}.\\

\textit{Correction Regarding At What Point Box-Range Predicates Are Included In the Predicate Covering:}
In \cref{subsubsec:producingCoveringOfBPrime}, previous descriptions suggested that box-range predicates
are added to entire coverings (forming $C'_F$ from $C_F$) shortly before formatting the 
results to show the user, as opposed to being included in the list of 
candidate predicates prior to any covering being formed.  
The pseudocode is uneffected and has been correct (in particular, algo. \ref{algo:getInitialListOfConditionsConsistentWithBoxes}); 
this is more evidence for the statement in \cref{footnote:methodology:pseudocodeMoreCorrectThanProseInCaseOfDiscrepency}.
As with the correction listed in \cref{subsubsec:appendix:errata:arxiv:d19M3y2021}, this erratum is purely
in regard to the description provided in this write-up, not any modification of the code.\footnote{
\label{footnote:erreta:willStateIfCodeEverHasToBeCorrect} Generally speaking, we will
explicitly state if any of the code has had to be modified to account for an item listed in the 
erreta. Short of any such explicit statement, one should assume that only descriptions are effected, not code.}  \\

\subsection{Errata for VMCAI 2022 Version of this Paper (\cite{DBLP:conf/vmcai/BayaniM22})}
\label{subsec:appendix:errata:VMCAI202}

\subsubsection{\dateUpdatePostVMCAITwentyTwentyTwo :} %
While not necessarily incorrect, the description of box-merging prior to
forming a predicate covering (the paragraph starting "Prior to proceeding, B may undergo some limited merging. [...]")
was placed in a location that seems likely to cause confusion. 
To be clear, the merging of any boxes occurs after the CEGAR-like analysis, not prior to 
pushing a box through the learned system.
This unfortunate placement of text
was the result of reshuffling 
into the main body of the write-up
content that had previously been in the appendix. 
Non-trivial modifications in the process of preparing the camera-ready
absorbed attention away from this high-level text placement issue.

The comment in \cref{subsubsec:appendix:errata:arxiv:d9M2y2022} regarding details of the 
covering process partially apply. Specifically, for the VMCAI write-up (\cite{DBLP:conf/vmcai/BayaniM22}),
$P''_b$ was missing.

\subsection{Errata for IJCAI XAI 2020  Version of this Paper \cite{IJCAIXAIFanoos2020}}
\label{subsec:appendix:errata:IJCAIXAI2021}

\subsubsection{\dateUpdatePostVMCAITwentyTwentyTwo : } %
Comments made in \cref{subsubsec:appendix:errata:arxiv:d19M3y2021} apply, as do 
those in \cref{subsubsec:appendix:errata:arxiv:d9M2y2022} regarding \cref{tab:questionDescriptions}
and the point at which box-range predicates are added.

\section{Further Materials}
\label{sec:appendix:furtherMaterials}

Further materials and pointers to them can be found in the file
\seqsplit{https://github.com/DBay-ani/FanoosFurtherMaterials/blob/master/manifest.xml} .
The size of this file is 18819 bytes, with a sha512 hash of
\seqsplit{3cd5a3015bb78bb0fd1d622779feff8263fd2ada587d090751193f0b56c157695b39ae705ab3c7b43ff20f162bcb69914b7e0c7e36a3e8f3cab2a3df5f5f3331}.

{\it \textbf{Update} \dateUpdatePostVMCAITwentyTwentyTwo:} As of this update to 
ArXiv, any changes to the additional material will be immutably recorded
via The Internet Archive at
\raggedright{\url{https://web.archive.org/web/*/https://github.com/DBay-ani/FanoosFurtherMaterials/blob/master/manifest.xml}}
as well as on Zenodo at \cite{david_bayani_2022_6069468}.
Either of these sources should be sufficient to confirm
the content of the manifest file, that file in turn providing identifying information
(file sizes and hashes, etc.) for the associated content.
This policy change helps
limit ArXiv updates to modifications/additions pertinent of this
document's principle content. ArXiv is not a tool intended to track entire project
portfolios, etc., and while we never violated that rule of thumb egregiously,
 hosting the metadata elsewhere\footnote{Noting, critically, that the
locations in question provide the content in an immutable fashion. In particular,
not even we, the authors, can modify the content once it is provided there.} puts us 
in better alignment with %
that notion. Further, this policy change 
will better indicate when
updates are made to this core description versus
when changes are made to supporting materials.

\end{document}